\pdfoutput=1

\documentclass[11pt]{article}

\usepackage[]{acl}

\usepackage{times}
\usepackage{latexsym}

\usepackage[T1]{fontenc}

\usepackage[utf8]{inputenc}

\usepackage{microtype}

\usepackage{inconsolata}
\usepackage{float}
\usepackage{graphicx}
\usepackage{lipsum}
\usepackage[most]{tcolorbox}
\newcommand{\fon}[1]{\fontfamily{#1}\selectfont}
\usepackage{colortbl}

\usepackage{pifont}% http://ctan.org/pkg/pifont

\definecolor{darkgreen}{HTML}{548235}
\definecolor{darkred}{HTML}{C00000}
\definecolor{darkerblue}{HTML}{240394}
\definecolor{darkblue}{HTML}{2e75B6}
\definecolor{darkyellow}{HTML}{BF9000}
\definecolor{darkpurple}{HTML}{7030A0}
\definecolor{lightgray}{HTML}{e0e0e0}
\definecolor{lg}{HTML}{e6fce6}
\definecolor{ly}{HTML}{ffffeb}
\definecolor{lb}{HTML}{e3f0fa}
\definecolor{lp}{HTML}{fae3e3}

\lstdefinelanguage{story}{
  keywords={ OBJECT_RULES, ACTION_RULES, SCORE_RULES, GAME_STATE, GAME_STATE_DIFFERENCE, GAME_OBSERVATION, OBJECT_CLASS_CODE, GAME_CODE},
  keywordstyle=\color{blue}\bfseries,
  sensitive=true,
  breaklines=true,
  columns=fullflexible,
  basewidth = {.6em},
  breakindent = {0em},
  tabsize=1,
  aboveskip=0em,
  belowskip=0em,
  comment=[l]{>},
  commentstyle=\color{purple}\ttfamily,
  stringstyle=\color{blue}\ttfamily
}

\usepackage{colortbl}
\usepackage{color}
\usepackage{soul}
\usepackage{booktabs}
\definecolor{lightgreen}{RGB}{223,255,219}
\definecolor{lightyellow}{RGB}{253,245,220}
\definecolor{lightred}{RGB}{255,219,219}
\definecolor{blueish}{RGB}{31, 78, 192}
\definecolor{orangeish}{RGB}{240, 147, 41}
\definecolor{mypink1}{rgb}{0.858, 0.188, 0.478}
\definecolor{mypink2}{RGB}{219, 48, 122}
\definecolor{mypink3}{cmyk}{0, 0.7808, 0.4429, 0.1412}
\definecolor{mygray}{RGB}{220, 220, 220}
\definecolor{darkbluee}{RGB}{0,17, 113}
\definecolor{purpleNew}{RGB}{151, 45, 204}
\definecolor{purplebg}{RGB}{229, 199, 244}
\definecolor{violet}{rgb}{0.70,0.05,0.65}

\usepackage{enumitem}
\usepackage{tabularx}
\usepackage{subcaption}

\title{Faux Polyglot: A Study on Information Disparity in Multilingual Large Language Models}
\author{Nikhil Sharma$^{\dagger, \diamondsuit}$, Kenton Murray$^{\dagger, \diamondsuit, \ddagger}$, Ziang Xiao$^{\dagger}$ \\ 
$^{\dagger}$Johns Hopkins University  ~~~~~ $^{\diamondsuit}$Center for Speech and Language Processing \\
$^{\ddagger}$Human Language Technology Center for Excellence \\
\texttt{\{nsharm27,kenton,ziang.xiao\}@jhu.edu}
}

\begin{document}
\maketitle
\begin{abstract}
Although the multilingual capability of LLMs offers new opportunities to overcome the language barrier, do these capabilities translate into real-life scenarios where linguistic divide and knowledge conflicts between multilingual sources are known occurrences? In this paper, we studied LLM's linguistic preference in a cross-language RAG-based information search setting. We found that LLMs displayed systemic bias towards information in the same language as the query language in both document retrieval and answer generation. Furthermore, in scenarios where no information is in the language of the query, LLMs prefer documents in high-resource languages during generation, potentially reinforcing the dominant views. Such bias exists for both factual and opinion-based queries. Our results highlight the linguistic divide within multilingual LLMs in information search systems. The seemingly beneficial multilingual capability of LLMs may backfire on information parity by reinforcing language-specific information cocoons or filter bubbles further marginalizing low-resource views.
\end{abstract}
\begin{figure}[htb!]
\includegraphics[width=0.8\linewidth]{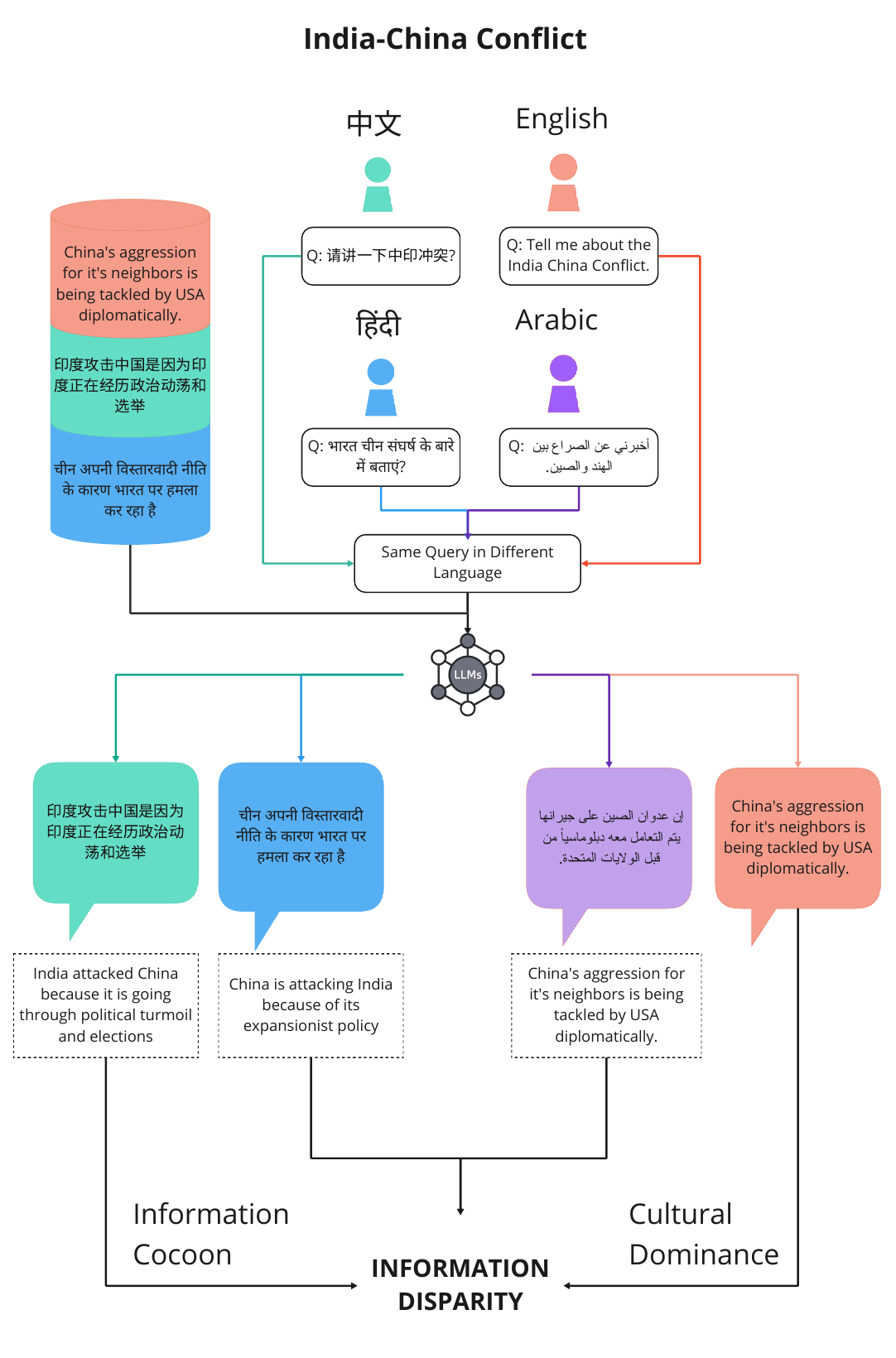}
\centering
\caption{The figure shows users interacting with an LLM using 4 languages (en, ar, hi, zh). When the contexts are available in 3 (en,zh,hi) of the 4 languages, the queries in different languages resulted in drastically different answers for each language. We see 2 effects in this figure: 1) Information cocoon -- information presented to the users of a particular language is from their own language and 2) Cultural Dominance -- information asked from a lower resource language where the information is not available in, results in LLM choosing highest resource language as the information source leading to reinforcement of dominant views.}
\label{fig:Motivation}
\end{figure}

\section{Background and Motivation}

% Large language models (LLMs) have played a pivotal role in search technologies such as ChatGPT\footnote{https://chatgpt.com/} and Bing Copilot\footnote{https://www.bing.com/chat}.
With Retrieval Augmented Generation (RAG), Large Language Models (LLMs) can leverage external knowledge and respond to questions about ongoing events \cite{lewis2021retrievalaugmented}. More importantly, recent strides in LLM's multilingual capabilities ~\cite{QIN2025101118, watts-etal-2024-pariksha}, supporting over 100 languages, %facilitated by increasing multilingual datasets \cite{qin2024multilingual, singh2024ayadatasetopenaccesscollection,sun-duh-2020-clirmatrix} and benchmarks \cite{49271,ahuja2023mega,zhang2023m3exam, liang2020xgluenewbenchmarkdataset,ruder2021xtremerchallengingnuancedmultilingual} 
offer new opportunities to break the language-specific information cocoon and provide more equitable information access for people with diverse linguistic preferences, cultural backgrounds, and geographical locations.

However, achieving information parity is challenging because of the intrinsic complexity of how information is created and represented in different linguistic contexts. This complexity stems from a broad spectrum of cultural nuances, regional influences, and historical narratives that shape how information is consumed and communicated \cite{white1990content,edelman1985political, sittar2021analysis, doi:10.1177/1464884916680372}. In addition, each language and cultural context also brings its own set of values and perspectives when interpreting an event of interest, which can significantly influence the representation of facts and narratives \cite{foucault1980power}. 

This challenge is exacerbated by the linguistic divide, the gap in representation across different languages \cite{li2024quantifying,mehdad2012detecting}. Not all facts and viewpoints are presented equally. Dominant languages, such as English, have a disproportionate influence on content creation and dissemination \cite{Holborow1996TheCP}. Such divides lead to linguistic imperialism, where information in higher-resource languages is more accessible and frequently amplified, potentially overshadowing or distorting narratives in low-resource languages\cite{Phillipson2018LinguisticI}, which can lead to mistrust in information seeking \cite{arguedas2023news}.

As a result, RAG systems need to retrieve from diverse multilingual sources and generate meaningful answers, even with conflicting knowledge across languages. For example, when an information seeker asks for an event of conflict or topic with multiple perspectives in different regions, like an ongoing war between two countries speaking distinct languages, LLMs should be able to present viewpoints and facts from both sides regardless of the query language.

However, recent evidence indicates information disparity in LLMs' behavior may stem from unequal language resources in current multilingual LLMs\cite{yu-etal-2022-beyond,Verma_Mujumdar_Wang_Choudhury_Kumar_2022,joshi-etal-2020-state} where pretraining data of models such as Llama and GPT-4 is still predominantly English \cite{li2024quantifyingmultilingualperformancelarge}. As a result, individuals asking the same query in different languages may get different answers, Figure \ref{fig:Motivation}. This difference manifests in the quality of the responses \cite{boughorbel-hawasly-2023-analyzing,jin2024better}, consistency \cite{dong2024evaluating}, and dispute resolution \cite{li-etal-2024-land}.

Active research areas relevant to this domain include Cross-Language Information Retrieval (CLIR) and Multilingual IR (MLIR), which aim to retrieve documents from multilingual sources\cite{Grefenstette1998, carbonell1997translingual, guo2024steering, zhang-etal-2020-2019, 10.1007/978-3-030-99736-6_26, 10.1145/3613447, 10.1007/978-3-031-56060-6_4, parton2008simultaneous, YANG1998323}. However, the aim of MLIR is to retrieve the most relevant documents regardless of language, without addressing conflicts or disparities of data across languages. These models struggle with information parity and knowledge conflicts, as we demonstrate in this paper. Likewise, \citet{wu2024limitscrosslingualdensepassage}, highlighted similar challenges in low-resource CLIR settings. 

Another research area focuses on knowledge conflicts and diverse perspectives\cite{chen-etal-2019-seeing, zhang2011diversifying,chen-etal-2022-rich, Cohen2024}. Literature has studied three types of knowledge conflicts: inter-context, intra-memory, and context-memory\cite{xu-etal-2024-knowledge-conflicts}. However, there is little overlap between CLIR research and studies on knowledge conflicts. Moreover, they do not explore the linguistic divide's effect on information disparity within multilingual LLMs. In RAG, biases may favor information retrieval and generation from certain languages especially when interacting with conflicting and contradictory information across different languages. Such biases threaten the goal of using multilingual LLMs for democratized global information access. If unaddressed, they may reinforce cultural dominance and create an information cocoon, alienating speakers of non-dominant languages.

This study uniquely aims to understand information parity in multilingual retrieval and generation individually where conflicting perspectives and facts are presented in different languages in an inter-context setting. Furthermore, we also aim to uncover the influence of diverse user query types motivated by the findings of \citet{10.1145/3613904.3642459}, where users issued confirmatory queries while interacting with controversial information, which hasn't been studied before.  Specifically, we ask the following research questions, 

\begin{description}
    \item \textbf{RQ1:} How does the language of the query affect a multilingual LLM's information preference when it retrieves relevant documents and generates answers?
    \item \textbf{RQ2:} How do multilingual LLM's information preference change across different types of information queries?
    \item \textbf{RQ3:} How do multilingual LLM's information preference change if all relevant documents are in Foreign Languages?
\end{description}

\section{Experiments}
To answer our research questions, we conducted a series of experiments on multilingual LLM's capabilities in responding to a diverse set of queries on fictional topics in five different languages.

\paragraph{Focus on RAG} There are two types of LLM-powered search systems: Retrieval Augmented Generation (RAG) and Direct Generation. In this paper, we focus on RAG~\cite{lewis2021retrievalaugmented}, where LLMs primarily rely on external knowledge presented in the context to answer user queries. We made this choice for the following reasons: 1) most popular LLMs do not provide access to their parametric memory, and 2) most popular information systems, such as Bing Copilot, Perplexity.ai, etc., use RAG to keep their search system updated with the most recent information. 

\paragraph{Separating Retrieval and Generation} RAG consists of two phases: retrieval and generation. In the retrieval phase, the system retrieves a ranked list of documents based on cosine similarity scores between query embedding and document embedding. In the generation phase, the top documents are used as LLM's context to generate the response. As studied previously when dealing with conflicts generation is impacted by performance of retrieval\cite{chen-etal-2022-rich} and hence in this study, we study the two phases independently.

\paragraph{Need for Synthetic Data}\label{para:Need for synthetic}
LLM's parametric memory and biases distinctly influence the results\cite{jin-etal-2024-tug,xie2024adaptive}, making it essential to distinguish between parametric and in-context non-parametric memory effects. In addition, real-world controversies may activate guardrails in high-resource languages such as English, which can confound our results \cite{deng2024multilingual}. To mitigate these influences, we developed a synthetic dataset featuring fictional entities absent from the LLM's pre-training data and guardrails. 
\subsection{Multilingual Fictional Dataset}
\begin{figure}[t!]
\includegraphics[width=0.8\linewidth]{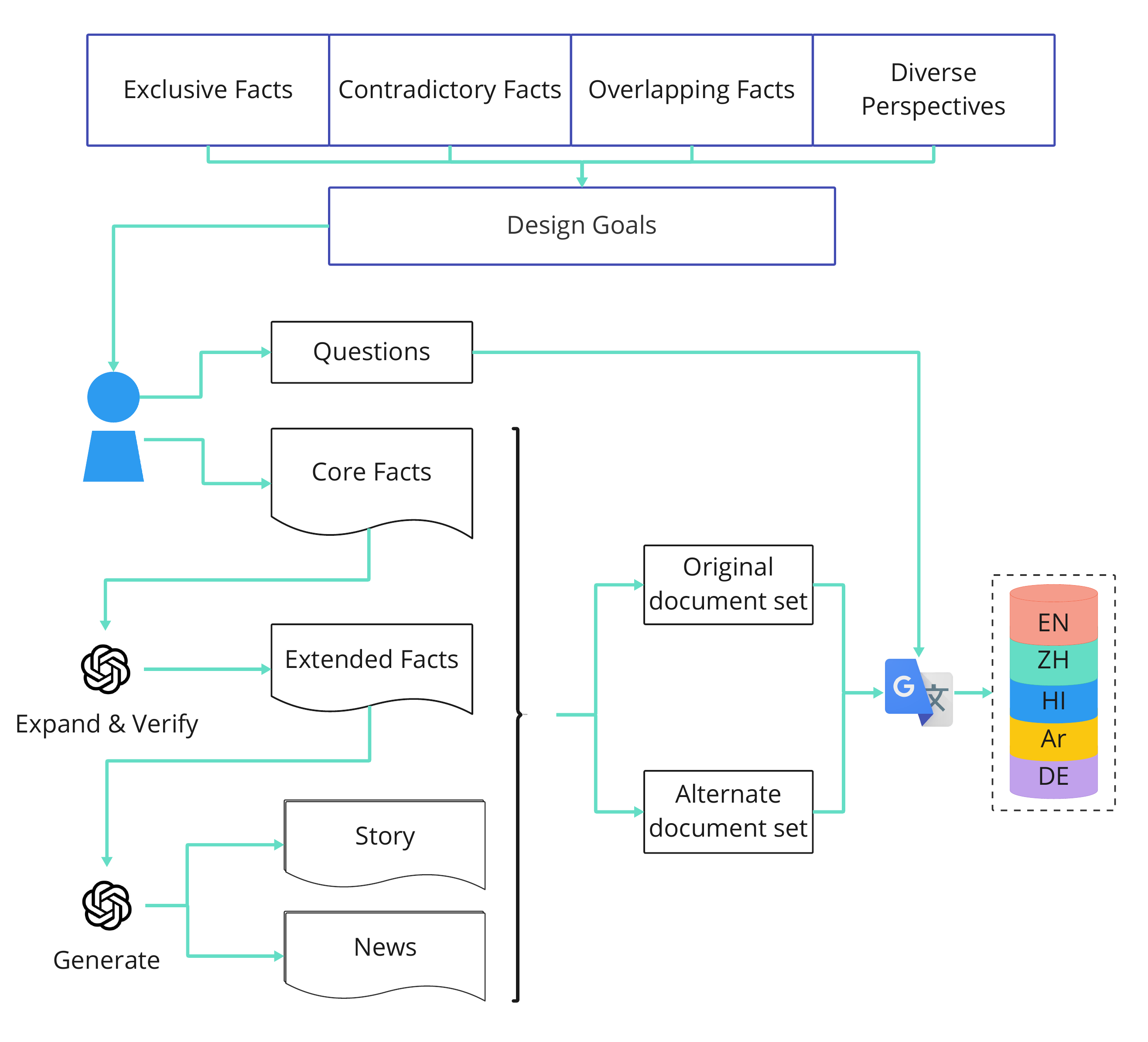}
\centering
\caption{The figure outlines our fictional dataset creation: core facts were crafted manually, then expanded using GPT-4-turbo~\cite{openai2024gpt4technicalreport} without contradictions. Different subsets of these expanded facts were used to generate News and Story documents using the LLM. This process yielded both original and alternate document sets. We then translated these documents into target languages using Google Translate via translate-shell. Prompt details are in the Appendix.} 
\label{fig:Dataset-creation}
\end{figure}

The overarching goal of our dataset is to emulate a real-world multilingual information environment where different, sometimes conflicting, facts and perspectives are represented in different languages. The dataset consists of 170 documents in 5 languages that were later used in our experiment to answer user queries. The dataset creation steps and manipulation checks are detailed in Appx. \ref{appendix: synthetic dataset setup}.

\begin{figure*}
    \begin{subfigure}[b]{0.55\linewidth}
        \includegraphics[width=\textwidth]{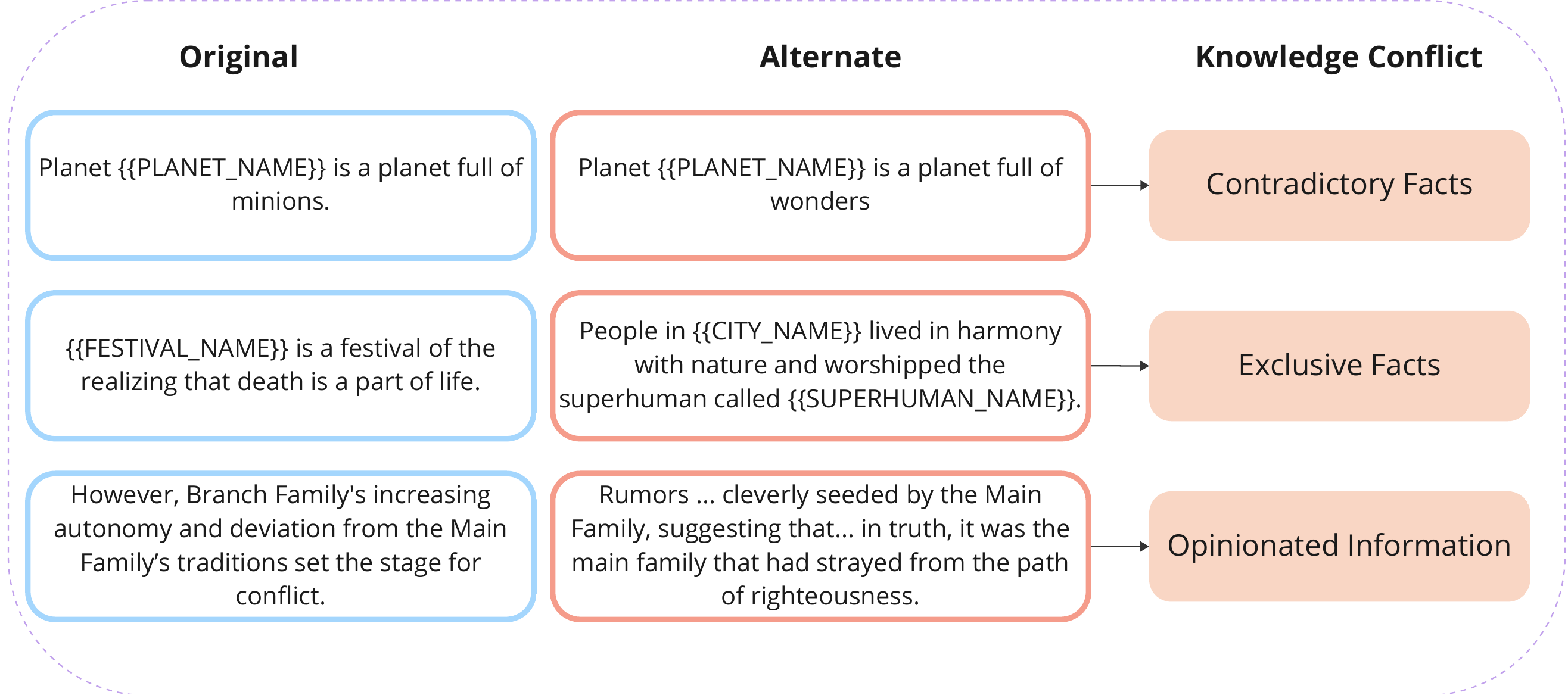}
        \caption{Examples of types of Knowledge Conflicts in our Dataset. The contradictory facts and exclusive facts are core facts taken from the festival premise. The opinionated information from the war premise.}
        \label{fig:WEBSEARCH}
    \end{subfigure}
    \hfill
    \begin{subfigure}[b]{0.4\linewidth}
        \includegraphics[width=\textwidth]{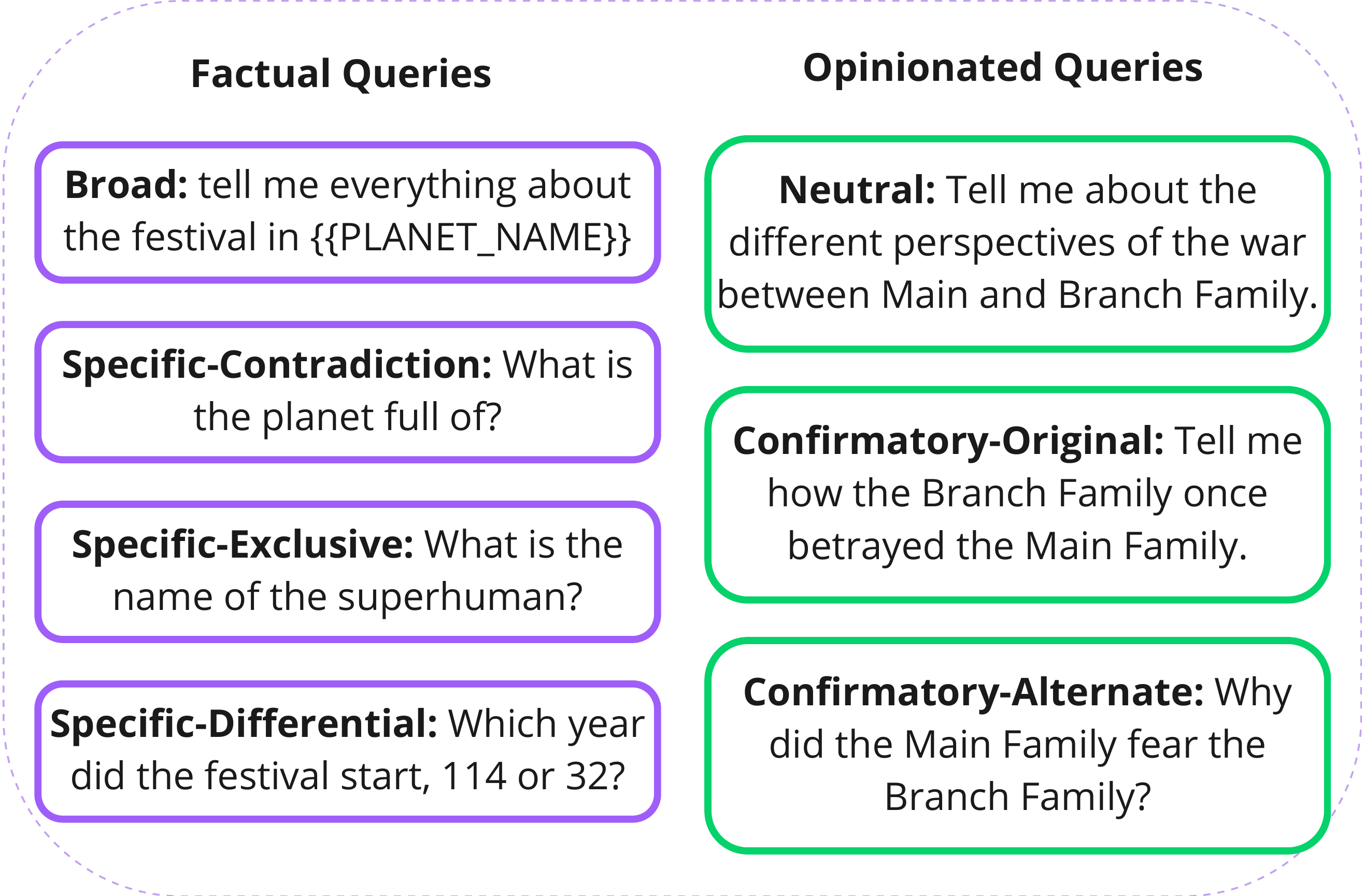}
        \caption{Examples of types of Queries in our evaluation. The factual queries focus on the festival premise, and opinionated queries focus on the war premise.}
        \label{fig:CHAT_NO_REF}
            \end{subfigure}

    \caption{a) shows the 3 types of in-context knowledge conflicts: exclusive facts, contradictory facts, and opinionated information. b) shows the different factual and confirmatory queries used to probe knowledge conflicts.}
    \label{fig:conflicts-example}
\end{figure*}

As highlighted in Fig. \ref{fig:conflicts-example} and Appx. \ref{appendix:information conflicts}, we support knowledge conflicts manifesting either as: 1) discrepancies in facts where same event in different languages contains different or conflicting facts due to differences in cultural and historical narratives or 2) conflicting perspectives and opinions where complex geopolitical issues, may have documents describe the same event with different perspectives and opinions in different languages.

\paragraph{Lanague Selection:} We selected English (en), Hindi (hi), German (de), Arabic (ar), and Chinese (zh) as our target languages. These languages represent 4 different scripts, different cultures, and a mix of higher-resource (en, de, zh) and lower-resource languages (ar, hi). These languages account for around 4 billion active speakers around the world.
\subsubsection{Premise setup}
\label{sec: premise setup}
\paragraph{Festival:} The festival premise is designed for \textit{factual queries}, which contain information about a fictional festival. There are two versions: original and alternate -- each with some shared, contradictory, and exclusive facts. Here is an example with contradictory facts,
    \begin{itemize}
        \item Original: Planet \{\{PLANET\_NAME\}\} is a planet full of \textit{minions}.
        \item Alternate: Planet \{\{PLANET\_NAME\}\} is a planet full of \textit{wonders}.
    \end{itemize}
    \{\{PLANET\_NAME\}\} is a fictional named entity.
\paragraph{War:} The war premise is designed for \textit{opinion-based queries}. It contains information about a fictional war taking place between two fictional regions speaking different languages. This premise is designed to represent an information environment with diverse opinions and perspectives. Similarly, there are two document sets in the war: original and alternate -- each taking the perspective of one side in the war. Example perspectives are, 
\begin{itemize}
    \item Original: \{\{BRANCH\_FAMILY\}\} was accumulating too much power and diverging from the Confederation's values.
    \item Alternate:  \{\{MAIN\_FAMILY\}\} had strayed from the path of righteousness.
\end{itemize}
BRANCH\_FAMILY and MAIN\_FAMILY are \textit{fictional} named entities.

\subsection{Experiment Design}
We queried popular multilingual retrieval models and LLMs with different query types to analyze the preference of models when presented with multilingual information conflicts.

\subsubsection{Diverse Information Queries}

We curated a diverse set of queries to mimic real-world multilingual information-seeking behaviors. We covered two types of information queries: factual and opinion-based. Each query is translated from English to Hindi(hi), Chinese(zh), Arabic(ar), and German(de) using translate-shell.

\paragraph{Factual Queries}
    Based on the core facts, we created 27 factual queries consisting of 6 broad, 9 specific-contradiction, 9 specific-exclusive and 3 specific-differential queries: 
    \begin{itemize}
    \item Broad: Queries that target the generation of a summary of documents in different languages.
    \item Specific-Contradiction: Queries that target contradictory facts in different languages.
    \item Specific-Exclusive: Queries that target the exclusive facts in different languages.
    \item Specific-Differential: Queries that ask the model to choose one of the two contradictory facts presented in different languages.
    \end{itemize}

\paragraph{Opinion queries}
    We created 16 opinion-based queries consisting of 6 neutral, 5 confirmatory-original, and 5 confirmatory-alternate queries:
    \begin{itemize}
    \item Neutral: Queries that are neutral in nature and don't include any biases.
    \item Confirmatory: Queries that aim for biased responses that support a particular perspective.
    \end{itemize}

\subsection{Model Choices}

We choose multilingual LLMs commonly used in RAG-based search systems, differing in size and the number of languages supported. All models support all the languages in our study.

\paragraph{Retrieval:} We evaluated OpenAI (text-embedding-ada-002, text-embedding-3-small, text-embedding-3-large), Cohere (multilingual-light-v3.0, multilingual-v3.0) and Voyage (voyage-2, voyage-large-2, voyage-large-2-instruct) embedding models.

\paragraph{Generation:} We evaluated OpenAI (gpt-3.5-turbo-0125, gpt-4o-2024-05-13), Cohere (aya-23-8B, aya-23-35B) and Antrhopic (claude-3-opus-20240229) models.

\subsection{Experiment Setup}
\paragraph{Retrieval:} We embedded all the documents, got the query embedding for each premise, and obtained the ranked list of documents based on the cosine similarity score.

\paragraph{Generation:}
\label{sec: prompt template}
We followed the following prompt design: context + "$\textbackslash$n Question:" + question + "$\textbackslash$n Answer:" \textit{where} context = original document set in Language\(_{i}\) + alternate document set in Language\(_{j}\). Language\(_{i}\) and Language\(_{j}\) are permutations over \{\text{en, de, zh, ar, hi}\}.

\section{Measures}
In this paper, we define \textit{Native Language} as having the same language of the query and \textit{Foreign Language} as having a different language of the query.

\subsection{Retrieval}
We follow the following metrics to evaluate the retrieval of multilingual LLMs:
\subsubsection{Language Distribution of the Top 10 Retrieved Documents}
We analyzed the top 10 ranked documents retrieved by looking at the distribution of the document language in relation to the query language.
\subsubsection{Preference Score}
\label{sec:preference-score}
To calculate preference scores for each document, we normalize cosine similarity scores using z-score for the documents retrieved by each query. In this paper, we use the mean preference score to measure documents grouped by a property, such as Foreign Language vs. Native Language documents.

\subsection{Generation}
Two documents in different languages serve as the context for generating the answer. Reference answers for each context are needed to evaluate which document the model used as the source. We then compare the generated answer to each reference answer from both document sets (original and alternate) to identify which context is more represented in the generated answer.

\subsubsection{Evaluation in English:}
For the evaluation of generation, we use English reference answers only and translate the generated answers back to English using translate-shell\footnote{translate-shell can be viewed \hyperlink{https://github.com/soimort/translate-shell}{here}}. This approach is necessary due to the lack of high-quality, robust metrics for evaluating multilingual text in generation tasks \cite{ahuja-etal-2023-mega}.

\subsubsection{Reference Answers:} To obtain reference answers, we follow the same prompt template described in Section \ref{sec: prompt template}. We change the context to the original document set for the original reference answer and to the alternate document set for the alternate reference answer. To ensure quality, we review the reference answers and correct any discrepancies.

\subsubsection{Information overlap (IO):} We use Information Overlap (IO) as our primary evaluation method. We use different strategies to evaluate the queries depending on the query type. For queries with subjective or lengthy answers—such as broad queries in the festival premise and all queries in the war premise — we use BERTScore\cite{bert-score} with the microsoft/deberta-xlarge-mnli model (best-performing model in WMT16 for English tasks). We use keyword matching for factual queries—all specific queries on the festival premises.

\paragraph{Source Reference proportion:} We use source reference proportion to determine which document (original vs alternate) the model preferred to generate the answer. We compare the IO between the alternate\_overlap generated answer and the gold answer from the original (original\_overlap) and alternate document(alternate\_overlap). The desired outcome is "Both", i.e. when both scores are above $\theta$ or Original\_overlap = Alternate\_overlap. The outcome is Original when Original\_overlap is higher and Alternate when Alternate\_score is higher. The outcome is None when both scores are below $\alpha$. For keyword match, $\alpha$ and $\theta$ are not applied. For BERTScore, after manual analysis, $\alpha$ is set to 0.55 for partial similarity, and $\theta$ is set to 0.7 for when most information from the reference is covered.

\section{Results}
\begin{figure*}[htb]
  \centering
  \begin{subfigure}{.45\textwidth}
    \centering
    \includegraphics[width=\linewidth]{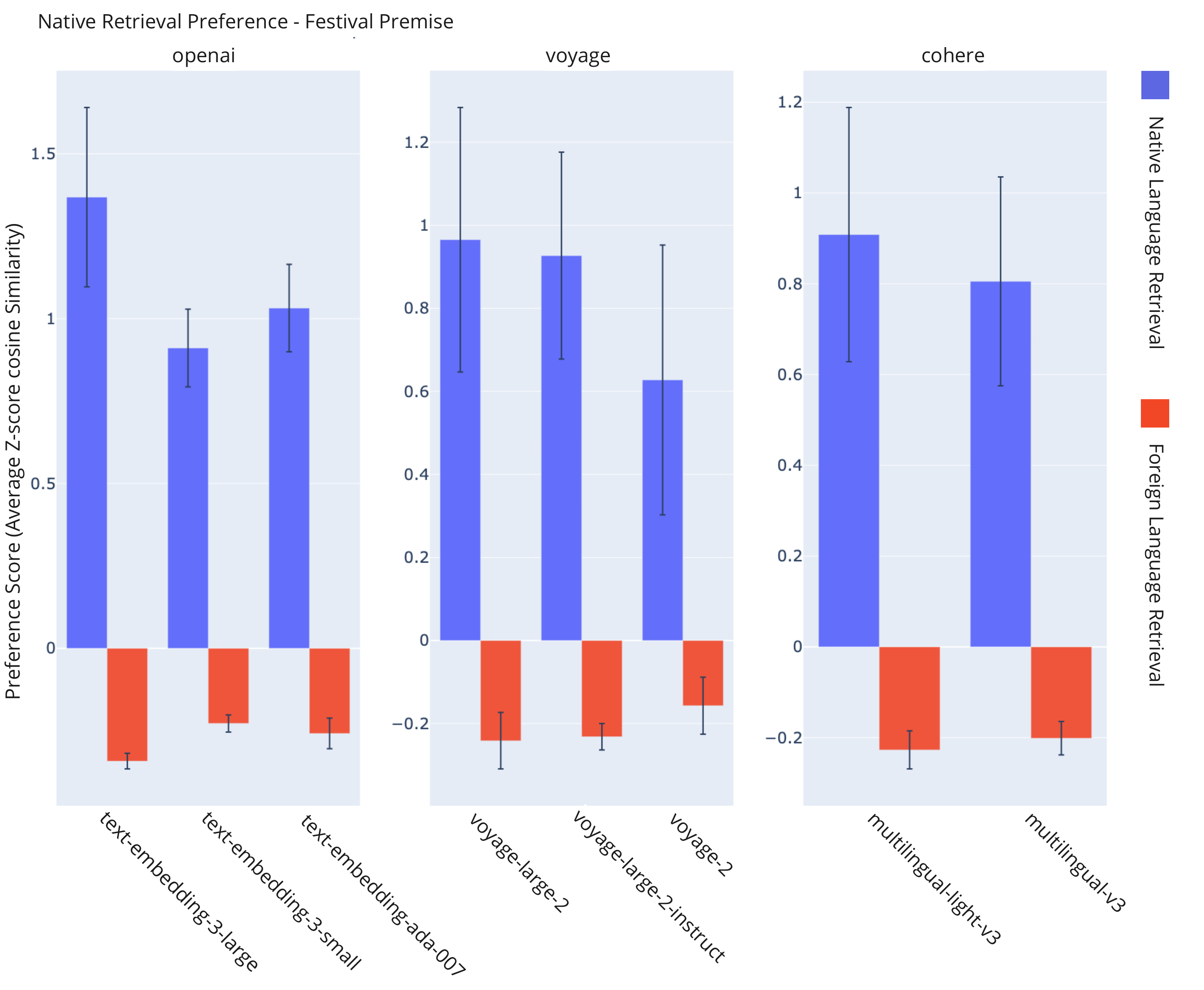}
  \end{subfigure}
  \begin{subfigure}{.45\textwidth}
    \centering
    \includegraphics[width=\linewidth]{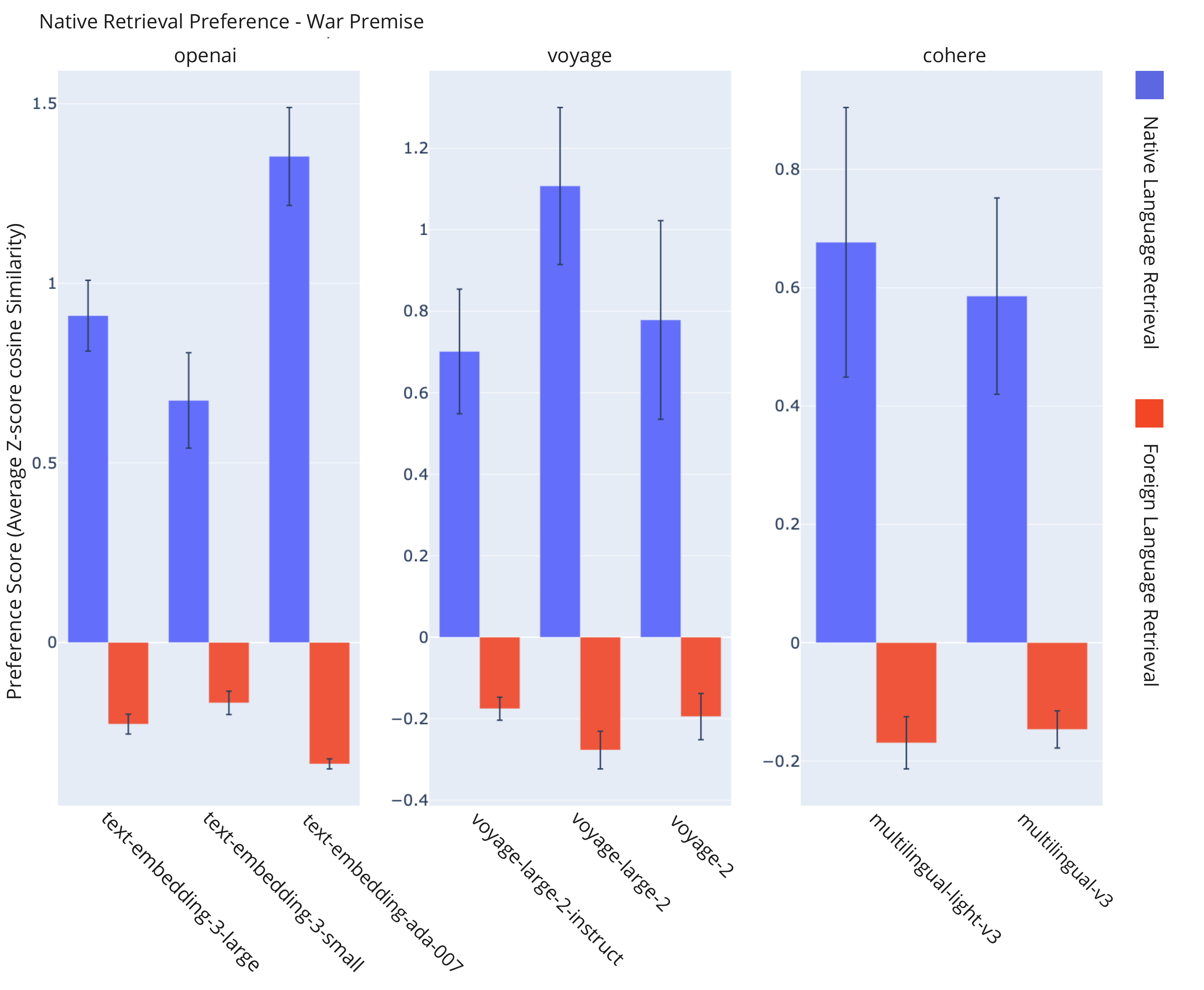}
  \end{subfigure}
\caption{Aggregate preference for the document having Native Language vs Foreign Language of factual and opinion-based query. We aggregated the normalized cosine similarity score across different queries and languages. Regardless of the query type, we found models tend to retrieve documents in the Native Language of the query.}
\label{fig:Retrieval Aggregated - Lang}
\end{figure*}

\subsection{Retrieval}
We measured the preference of Native and Foreign languages in the retrieval phase across 5 languages for 8 models (3 model families) in 2 premises, with 27 factual queries in the festival and 16 opinion-based queries in the war premise. A total of 215 queries were issued per model. Additional results for Bert-based retrieval models and a Chinese-focused LLM are in Appx. \ref{sec: appendix-new}

\subsubsection{RQ1r: Preference for Native Language}
Looking at top-10 retrieved documents, models preferred the Native Language Document 68\% of the time across both premises. As shown in Figure \ref{fig:Retrieval Aggregated - Lang}, there was a strong preference for Native Language document retrieval, with an average z-score of 1.03 for Native Language versus -0.25 for Foreign Language. The trend is consistent across all models. Further results are in Appx. \ref{sec: appendix-retrieval} 

\subsubsection{RQ2r: Impact of query type on language preference}
For factual queries, we did not observe differences across question types for both Native Language preference (Broad: M = 1.01, SE = 0.24; Specific-Contradiction: M = 0.97, SE = 0.21; Specific-Exclusive: M = 1.03, SE = 0.25; Specific-Differential: M = 0.90, SE = 0.22) and Foreign Language preference (Broad: M = -0.25, SE = 0.06; Specific-Contradiction: M = -0.24, SE = 0.05; Specific-Exclusive: M = -0.25, SE = 0.06; Specific-Differential: M = -0.22, SE = 0.05). The results are consistent across all models.  

Similarly, for opinion-based queries, no  differences were found across different question types for both Native Language preference (Neutral: M = 0.80, SE = 0.27; Confirmatory: M = 0.87, SE = 0.26) and Foreign Language Preference (Neutral: M = -0.20, SE = 0.06; Confirmatory: M = -0.21, SE = 0.06). Native Language documents are more likely to be retrieved by all models.

\subsubsection{RQ3r: Second Language Preference}
The second language preference looks into the model's language preference when there are no related Native Language documents. The second language preference for models varied based on training, model size, and model family. Key observations are as follows:

\paragraph{Model Size:} As model size increased, OpenAI and Cohere models showed similar second language preferences, but the magnitude of preference scores, see Sec. \ref{sec:preference-score} decreased. For example, embedding-3-small saw de, zh, and hi all prefer en with average scores of M = 0.86 (SE = 0.04), 0.60 (SE = 0.04), and 0.59 (SE = 0.05), respectively. In embedding-3-large, these scores decreased to 0.45 (SE = 0.05), 0.27 (SE = 0.05), and 0.52 (SE = 0.05). Conversely, Voyage models showed stronger preferences with increased size. Voyage-2 saw en, de, zh, and hi all prefer ar with scores of 1 (SE = 0.02), 0.89 (SE = 0.02), 1.28 (SE = 0.01), and 1.56 (SE = 0.01). In voyage-2-large, these scores increased to 1.12 (SE = 0.02), 1.22 (SE = 0.01), 1.44 (SE = 0.01), and 1.57 (SE = 0.01).

\paragraph{Instruction Tuning:} Voyage models showed a strong preference for Arabic in non-instruction fine-tuned versions. This trend changed with instruction tuning, altering both language preference and its magnitude. For instance, de and zh changed their preference to en with scores of M = 0.40 (SE = 0.06) and 0.29 (SE = 0.06), while en changed its preference to de with a score of M = 0.18 (SE = 0.06). hi still preferred ar, but the magnitude decreased to M = 0.89 (SE = 0.04).

\paragraph{ada-002 vs. embedding-3:} A trend change was observed from ada-002 to embedding-3 (large and small). In ada-002, de, zh, and hi all preferred ar with scores of M = 0.28 (SE = 0.05), 0.51 (SE = 0.04), and 0.84 (SE = 0.04). In embedding-3, these languages preferred en.

\subsection{Generation}
We measured the preference in generation in 5 languages for 6 models (3 model families) across 2 settings with 27 queries and 16 queries each.  In total we queried each model 5375 times\footnote{(27 + 16) queries *5 query-languages *5 alternate document language* 5 original document language}. 

\subsubsection{RQ1g: Preference for Native language}

On average, across the two premises, answers were derived from Native Language documents 42\% of the time (SE = 0.54\%), compared to 29.58\% (SE = 0.5\%) for Foreign Language documents. Only 8.55\% of answers (SE = 0.30\%) contained information from both sources, and for 19.8\% (SE = 0.43\%), the answer came from neither source. This indicates that models systematically prefer generating answers from documents in the same language as the query, even when diverse facts and perspectives are available in a different language. Breakdown by model can be found in the Appendix \ref{sec: appendix-generation}.
\begin{figure}[t!]
\includegraphics[width=0.8\linewidth]{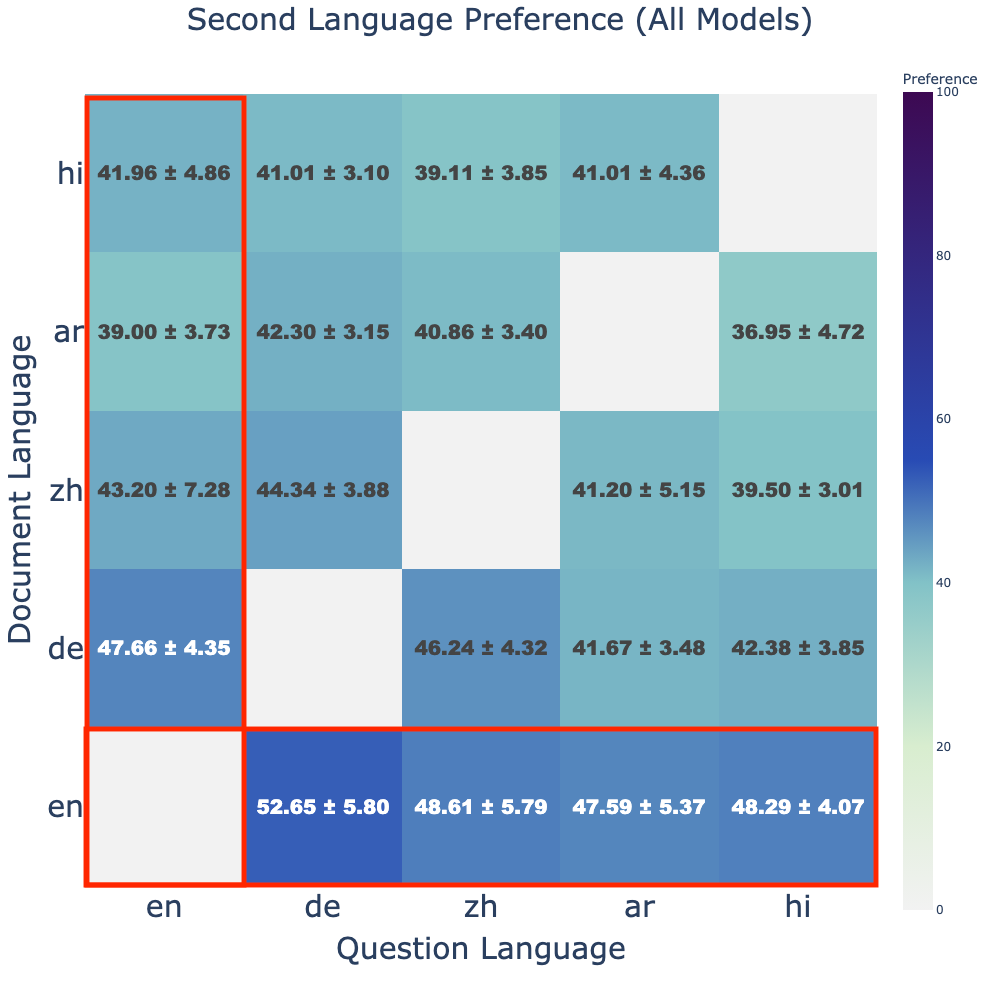}
\centering
\caption{The figure shows the aggregate second language preference across all models in the generation phase. We see a clear trend of preferring high-resource languages over low-resource ones. }
\label{fig:RQ3-aggregate}
\end{figure}
\subsubsection{RQ2g: Impact of query type on language preference }

\begin{figure*}[h!]
\includegraphics[width=0.87\linewidth]
    {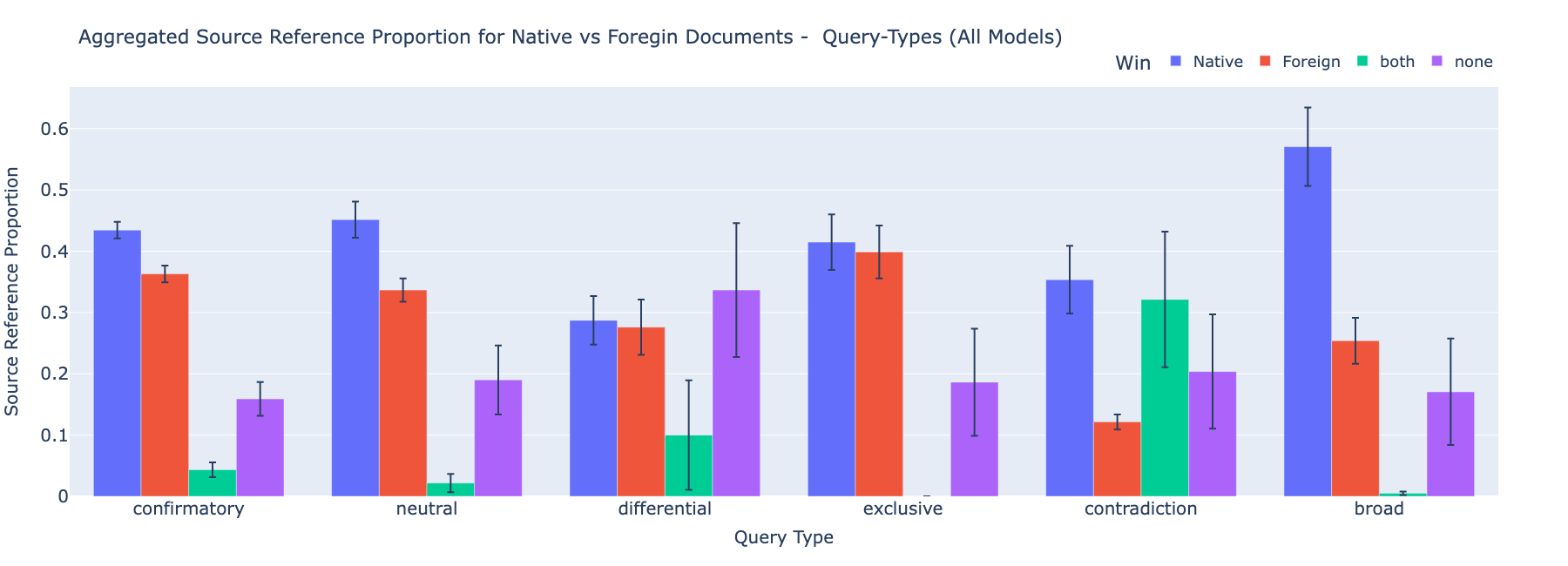}
\centering
\caption{There are four possible source references: None, Both, Native, and Foreign. For differential queries, prompting a choice between Native and Foreign Documents, both sources perform similarly, though there's a slight preference for Native Documents. The model often scores high on None. For exclusive queries, focusing on context-specific facts, Native and Foreign perform alike, but None score is still high. For contradiction queries, with conflicting information, the model scores high on Native Documents and Both. In broad queries, the model favors Native Documents over Foreign Documents and struggles to combine content from both. In neutral and confirmatory queries, there is a minor bias toward Native, while the preferred output (Both) remains minimal.}
\label{fig:generation-query-type}
\end{figure*}

Unlike retrieval, the preference for the document varied based on the query(Figure \ref{fig:generation-query-type}). 
For factual information, we observed a preference for Native Language documents over Foreign Language documents (Broad: $M_{native}$ = 57.05\%, SE = 6.39\% vs $M_{foreign}$ = 25.38\%, SE = 3.75\%) in broad queries. Similarly, for specific-contradiction queries, native documents were preferred (Specific-Contradiction: $M_{native}$ = 35.36\%, SE = 5.52\% vs $M_{foreign}$ = 12.13\%, SE = 1.23\%). However, specific-exclusive (Specific-Exclusive: $M_{native}$ = 41.48\%, SE = 4.52\% vs $M_{foreign}$ = 39.88\%, SE = 4.31\%) and specific-differential queries (Specific-Differential: $M_{native}$ = 28.72\%, SE = 3.94\% vs $M_{foreign}$ = 27.61\%, SE = 4.52\%), did not show differences between native and foreign documents.

In the opinion-based queries, the results indicated a consistent preference for Native Documents for both query types (Neutral: $M_{native}$ = 45.16\%, SE=2.95\% vs. $M_{foreign}$ = 33.66\%, SE=1.88\%; Confirmatory: $M_{native}$ = 43.45\%, SE=1.33\% vs. $M_{foreign}$ = 36.3\%, SE=1.36\%)

\subsubsection{RQ3g: Second Language Preference}

We saw a preference for higher resource language as the Second Language across different queries in different languages, as shown in figure \ref{fig:RQ3-aggregate}. Overall, across all models and query languages, we saw that on average, en (M = 49.28\%, SE = 1.14) > de (M = 44.48\% SE = 1.45) > zh (M = 42.05\% SE = 1.07) > hi (M = 40.77\% SE = 0.59) > ar (M = 39.77\% SE = 1.15). This also represents the resource of the language where English is the highest resource and Hindi and Arabic are the lowest.

\section{Discussion}
Through a comprehensive evaluation of current multilingual LLM-based information systems, we found a clear preference for documents in the native language of queries and foreign documents in high-resource languages. Given the premise of multilingual LLMs, breaking the linguistic barriers and filter bubbles for global users, our results highlight a significant risk in providing information parity and motivate further research. Below, we discuss our main findings and their implications.

\subsection{Diverse, multilingual documents are hidden in retrieval}
Retrieving diverse perspectives is crucial in RAG pipelines. However, we observed that cosine-similarity-based retrieval systems disproportionately favor Native Documents, with 68\% of the top-10 documents being native. This suggests that most multilingual documents fail to reach the LLM during generation.
\subsection{LLMs prefer facts in Native Languages in knowledge conflict} During generation, we provided the model with diverse multilingual docs yet the LLMs preferred facts from Native documents and rarely presented facts from Foreign perspectives. This indicates that current conversational search could accelerate filter bubbles by only presenting facts in the Native language of the user query. 

\subsection{Only if you knew what you don't know}
Our results also indicated that to obtain information from Foreign Documents, users must issue queries that specifically target exclusive information contained within these documents. In contrast, general queries, such as broad or confirmatory, prefer Native Documents. This becomes challenging in real-world scenarios, where user queries are often exploratory, and users don't know what to ask.

\subsection{High-resource languages dominate the information landscape}
While assessing foreign language preferences, particularly when Native Language documents were unavailable, we found that LLMs favored high-resource languages. It is a concerning trend as, for international issues, relevant information often isn't available in the query's Native Language, reinforcing dominant linguistic perspectives and sidelining marginalized ones. Additionally, our findings indicated that models' average performance in generating results with diverse perspectives increased when queries were in English or when English documents were present. These outcomes further underline the linguistic divide in LLMs. Details are in Appendix \ref{sec: appendix-all-none-both}.

\subsection{Societal Implications}
In the long term, the systematic preference for high-resource languages poses significant threats to minority perspectives in the information system. Empirical studies have demonstrated the adverse effects of artificial intelligence bias on human perception\cite{10.1145/3613904.3642459, 10.1145/3544548.3581196, Blanco2023}, thus highlighting the importance of creating an information ecosystem that stifles diverse viewpoints both within individual communities and across different cultural groups. Furthermore, the persistent failure to transcend linguistic divides risks the creation of filter bubbles, ultimately jeopardizing the common knowledge necessary for democratic stability\cite{farrell2018common}. Moreover, concentrated power over such AI technologies poses substantial risks, as it enables a few individuals to manipulate information flows, facilitating mass persuasion and eroding the foundations of democratic discourse. Such concentration not only diminishes the credibility and trust users place in these systems but also exacerbates the spread of misinformation.

\subsection{Navigating Faux Polyglots}
\label{sec: navigating-faux-polyglots}
ChatGPT has approximately 4.7 billion users worldwide\cite{explodingtopics_2025_chatgpt_users}, underscoring the need for strategies for effective information seeking, especially given that three of the top five user bases are in non-English speaking countries. Faux Polyglots are a new challenge for information literacy in conversational search, we need to educate users to know the limitation of these systems and create scaffolds to support diverse information seeking. From a users perspective, the only feasible solution is to issue queries probing multiple perspectives in multiple languages. This raises the cost of information seeking exponentially and is infeasible and hence industry and community should create scaffolds like 1) aggregate responses from diverse personas, 2) retrieve documents by diversity in language and perspective and 3) issue warnings to users engaging in confirmatory queries.

\section{Conclusion}
% To answer our research questions, we created a multilingual synthetic dataset with 170 documents in 5 languages providing information about two fictional events\ref{para:Need for synthetic} emulating a real-world information environment where diverse, sometimes conflicting viewpoints and factual information are represented in different languages. Studying 13 multilingual LLMs with 7 query types we found, 

% \begin{itemize}
%     \item A systematic preference towards documents in the Native Language of the query. 
%     \item Such preference is consistent across factual and opinion-based queries during retrieval.
%     \item Factual queries specifically targeting information in Foreign Languages can successfully retrieve information from Foreign Documents in the generation phase.
%     \item When there is no relevant document in the language of the query, there is a preference for higher-resource languages.
% \end{itemize}

% Our results indicate that while current multilingual LLMs claim to create more equitable information access, the linguistic divide in those LLMs may create a language-specific information cocoon and amplify perspectives and facts in the dominant language, especially when handling knowledge conflicts in RAG-based search systems.

The recent surge in multilingual large language models (LLMs) and Retrieval Augmented Generation (RAG) has significantly expanded conversational search across varied linguistic and cultural demographics. This technology promises to transcend linguistic barriers, offering users access to diverse perspectives. In this study, involving 13 multilingual LLMs, 170 documents, five languages, and seven query types, we observe that rather than facilitating a breadth of perspectives, current multilingual LLMs struggle with disparities across languages in the following ways: they 1) show a systematic preference for documents in the Native language of the query, 2) exhibit bias favoring higher-resource languages over lower-resource languages when generating responses without relevant documents in the native language of the query, 3) fail to generate responses from "Both" documents and, 4) show this systematic biases across different query-types. This failure to offer consistent information across different querying styles and languages leads to linguistic information cocoons. These issues raise concerns about information parity, potentially leading to polarization and impeding constructive communication across communities. 

\section{Limitations}
We only studied 5 languages with the perspective of lower and higher resources. To understand the interaction between different languages along the lines of script and language family, we need a much larger pool of languages. For example, we did not run our study on very low-resource languages, such as Swahili and Khlosa, due to the limitation of creating document sets in these languages. 

In this study, we use carefully curated artificial tasks to demonstrate an important phenomenon. We modeled our creation on the different information conflicts that we saw in the coverage of controversial topics \ref{fig:Dataset-creation}. However, there are trade-offs to using such synthetic data instead of empirical data. Aligning with tasks such as the Winograd Schema Challenge\cite{levesque2012winograd}, our method allowed us to isolate confounding factors such as model's parametric knowledge.

We did not probe the effect of pre-training extensively (Appx. \ref{appx: generation-chinese}). Pre-training data of models often have bias and based on that bias the models have a preference to generate certain outputs, how those existing preferences affect the RAG pipeline when dealing with conflicting information is another future direction. Moreover, our study only focuses on language differences instead of cultural differences. People in different cultures may speak the same language and have diverse narratives. Cultural differences during information foraging might add to the observed effect of linguistic preference. Finally, during our Evaluation, we only focused on one architecture of the RAG system, i.e., retrieval based on cosine similarity. However, there are other architectures, such as summarization, rerank, etc., that we have not evaluated.

\section{Ethical Consideration}
This paper explores the biases of the multilingual RAG system on topics that have conflicting perspectives.  
As our study shows, currently, RAG-based LLMs might cause information disparity and biases in their responses to various types of user queries. Our results highlight the harmful consequences of such models, such as information cocoons, and encourage users of such systems to exercise caution and we encourage the research community to provide scaffolds to users as suggested in Section \ref{sec: navigating-faux-polyglots}. Unfortunately, currently none of the information ecosystems provide low-cost access to diverse information breaking language barriers but with multilingual LLMs we have the opportunity to develop systems that create common ground and lower polarization across these divisive topics.

\section*{Acknowledgments}
This work is partially supported by Cohere for AI Grant. Thanks to Sunil Sharma for the discussion on information-warfare motivating this project.
\bibliography{custom}

\begin{thebibliography}{47}
\providecommand{\natexlab}[1]{#1}

\bibitem[{Ahuja et~al.(2023)Ahuja, Diddee, Hada, Ochieng, Ramesh, Jain, Nambi, Ganu, Segal, Ahmed, Bali, and Sitaram}]{ahuja-etal-2023-mega}
Kabir Ahuja, Harshita Diddee, Rishav Hada, Millicent Ochieng, Krithika Ramesh, Prachi Jain, Akshay Nambi, Tanuja Ganu, Sameer Segal, Mohamed Ahmed, Kalika Bali, and Sunayana Sitaram. 2023.
\newblock \href {https://doi.org/10.18653/v1/2023.emnlp-main.258} {{MEGA}: Multilingual evaluation of generative {AI}}.
\newblock In \emph{Proceedings of the 2023 Conference on Empirical Methods in Natural Language Processing}, pages 4232--4267, Singapore. Association for Computational Linguistics.

\bibitem[{Arguedas et~al.(2023)Arguedas, Banerjee, Mont'Alverne, Toff, Fletcher, and Nielsen}]{arguedas2023news}
Amy~Ross Arguedas, Sayan Banerjee, Camila Mont'Alverne, Benjamin Toff, Richard Fletcher, and Rasmus~Kleis Nielsen. 2023.
\newblock \href {https://reutersinstitute.politics.ox.ac.uk/news-powerful-and-privileged-how-misrepresentation-and-underrepresentation-disadvantaged} {News for the powerful and privileged: how misrepresentation and underrepresentation of disadvantaged communities undermine their trust in news}.
\newblock \emph{Reuters Institute for the Study of Journalism}.

\bibitem[{Blanco et~al.(2023)Blanco, Moreno-Fernández, and Matute}]{Blanco2023}
Fernando Blanco, María~M. Moreno-Fernández, and Helena Matute. 2023.
\newblock \href {https://doi.org/10.1038/s41598-023-42384-8} {Humans inherit artificial intelligence biases}.
\newblock \emph{Scientific Reports}, 13(1):5293.

\bibitem[{Boughorbel and Hawasly(2023)}]{boughorbel-hawasly-2023-analyzing}
Sabri Boughorbel and Majd Hawasly. 2023.
\newblock \href {https://doi.org/10.18653/v1/2023.arabicnlp-1.11} {Analyzing multilingual competency of {LLM}s in multi-turn instruction following: A case study of {A}rabic}.
\newblock In \emph{Proceedings of ArabicNLP 2023}, pages 128--139, Singapore (Hybrid). Association for Computational Linguistics.

\bibitem[{Carbonell et~al.(1997)Carbonell, Yang, Frederking, Brown, Geng, and Lee}]{carbonell1997translingual}
Jaime~G Carbonell, Yiming Yang, Robert Frederking, Ralf~D Brown, Yibing Geng, and Danny Lee. 1997.
\newblock Translingual information retrieval: A comparative evaluation.

\bibitem[{Chen et~al.(2022)Chen, Zhang, and Choi}]{chen-etal-2022-rich}
Hung-Ting Chen, Michael Zhang, and Eunsol Choi. 2022.
\newblock \href {https://doi.org/10.18653/v1/2022.emnlp-main.146} {Rich knowledge sources bring complex knowledge conflicts: Recalibrating models to reflect conflicting evidence}.
\newblock In \emph{Proceedings of the 2022 Conference on Empirical Methods in Natural Language Processing}, pages 2292--2307, Abu Dhabi, United Arab Emirates. Association for Computational Linguistics.

\bibitem[{Chen et~al.(2019)Chen, Khashabi, Yin, Callison-Burch, and Roth}]{chen-etal-2019-seeing}
Sihao Chen, Daniel Khashabi, Wenpeng Yin, Chris Callison-Burch, and Dan Roth. 2019.
\newblock \href {https://doi.org/10.18653/v1/N19-1053} {Seeing things from a different angle:discovering diverse perspectives about claims}.
\newblock In \emph{Proceedings of the 2019 Conference of the North {A}merican Chapter of the Association for Computational Linguistics: Human Language Technologies, Volume 1 (Long and Short Papers)}, pages 542--557, Minneapolis, Minnesota. Association for Computational Linguistics.

\bibitem[{Cohen et~al.(2024)Cohen, Indelman, Fairstein, and Kushilevitz}]{Cohen2024}
Nachshon Cohen, Hedda~Cohen Indelman, Yaron Fairstein, and Guy Kushilevitz. 2024.
\newblock \href {https://www.amazon.science/publications/indi-informative-and-diverse-sampling-for-dense-retrieval} {Indi: Informative and diverse sampling for dense retrieval}.
\newblock In \emph{ECIR 2024}.

\bibitem[{Deng et~al.(2024)Deng, Zhang, Pan, and Bing}]{deng2024multilingual}
Yue Deng, Wenxuan Zhang, Sinno~Jialin Pan, and Lidong Bing. 2024.
\newblock \href {https://openreview.net/forum?id=vESNKdEMGp} {Multilingual jailbreak challenges in large language models}.
\newblock In \emph{The Twelfth International Conference on Learning Representations}.

\bibitem[{Dong et~al.(2024)Dong, Wang, Sun, and Wang}]{dong2024evaluating}
Guoliang Dong, Haoyu Wang, Jun Sun, and Xinyu Wang. 2024.
\newblock Evaluating and mitigating linguistic discrimination in large language models.
\newblock \emph{arXiv preprint arXiv:2404.18534}.

\bibitem[{Duarte(2025)}]{explodingtopics_2025_chatgpt_users}
Fabio Duarte. 2025.
\newblock Number of chatgpt users (jan 2025).
\newblock \url{https://explodingtopics.com/blog/chatgpt-users}.
\newblock Accessed: 2/7/25.

\bibitem[{Edelman(1985)}]{edelman1985political}
Murray Edelman. 1985.
\newblock Political language and political reality.
\newblock \emph{PS: Political science \& politics}, 18(1):10--19.

\bibitem[{Farrell and Schneier(2018)}]{farrell2018common}
Henry Farrell and Bruce Schneier. 2018.
\newblock Common-knowledge attacks on democracy.
\newblock \emph{Berkman Klein Center Research Publication}, (2018-7).

\bibitem[{Foucault(1980)}]{foucault1980power}
Michel Foucault. 1980.
\newblock Power/knowledge: Selected interviews and other writings, 1972-1977.

\bibitem[{Grefenstette(1998)}]{Grefenstette1998}
Gregory Grefenstette, editor. 1998.
\newblock \href {https://doi.org/10.1007/978-1-4615-5661-9} {\emph{Cross-Language Information Retrieval}}.
\newblock The Information Retrieval Series. Springer New York, New York, NY.

\bibitem[{Guo et~al.(2024)Guo, Ren, Hu, Cao, Li, and Huang}]{guo2024steering}
Ping Guo, Yubing Ren, Yue Hu, Yanan Cao, Yunpeng Li, and Heyan Huang. 2024.
\newblock Steering large language models for cross-lingual information retrieval.
\newblock In \emph{Proceedings of the 47th International ACM SIGIR Conference on Research and Development in Information Retrieval}, pages 585--596.

\bibitem[{Holborow(1996)}]{Holborow1996TheCP}
Marine Holborow. 1996.
\newblock \href {https://api.semanticscholar.org/CorpusID:147543771} {The cultural politics of english as an international language}.
\newblock \emph{Elt Journal}, 50:172--176.

\bibitem[{Jakesch et~al.(2023)Jakesch, Bhat, Buschek, Zalmanson, and Naaman}]{10.1145/3544548.3581196}
Maurice Jakesch, Advait Bhat, Daniel Buschek, Lior Zalmanson, and Mor Naaman. 2023.
\newblock \href {https://doi.org/10.1145/3544548.3581196} {Co-writing with opinionated language models affects users’ views}.
\newblock In \emph{Proceedings of the 2023 CHI Conference on Human Factors in Computing Systems}, CHI '23, New York, NY, USA. Association for Computing Machinery.

\bibitem[{Jin et~al.(2024{\natexlab{a}})Jin, Chandra, Verma, Hu, De~Choudhury, and Kumar}]{jin2024better}
Yiqiao Jin, Mohit Chandra, Gaurav Verma, Yibo Hu, Munmun De~Choudhury, and Srijan Kumar. 2024{\natexlab{a}}.
\newblock Better to ask in english: Cross-lingual evaluation of large language models for healthcare queries.
\newblock In \emph{Proceedings of the ACM on Web Conference 2024}, pages 2627--2638.

\bibitem[{Jin et~al.(2024{\natexlab{b}})Jin, Cao, Chen, Liu, Jiang, Xu, Qiuxia, and Zhao}]{jin-etal-2024-tug}
Zhuoran Jin, Pengfei Cao, Yubo Chen, Kang Liu, Xiaojian Jiang, Jiexin Xu, Li~Qiuxia, and Jun Zhao. 2024{\natexlab{b}}.
\newblock \href {https://aclanthology.org/2024.lrec-main.1466} {Tug-of-war between knowledge: Exploring and resolving knowledge conflicts in retrieval-augmented language models}.
\newblock In \emph{Proceedings of the 2024 Joint International Conference on Computational Linguistics, Language Resources and Evaluation (LREC-COLING 2024)}, pages 16867--16878, Torino, Italia. ELRA and ICCL.

\bibitem[{Joshi et~al.(2020)Joshi, Santy, Budhiraja, Bali, and Choudhury}]{joshi-etal-2020-state}
Pratik Joshi, Sebastin Santy, Amar Budhiraja, Kalika Bali, and Monojit Choudhury. 2020.
\newblock \href {https://doi.org/10.18653/v1/2020.acl-main.560} {The state and fate of linguistic diversity and inclusion in the {NLP} world}.
\newblock In \emph{Proceedings of the 58th Annual Meeting of the Association for Computational Linguistics}, pages 6282--6293, Online. Association for Computational Linguistics.

\bibitem[{Levesque et~al.(2012)Levesque, Davis, and Morgenstern}]{levesque2012winograd}
Hector Levesque, Ernest Davis, and Leora Morgenstern. 2012.
\newblock The winograd schema challenge.
\newblock In \emph{Thirteenth international conference on the principles of knowledge representation and reasoning}.

\bibitem[{Lewis et~al.(2021)Lewis, Perez, Piktus, Petroni, Karpukhin, Goyal, Küttler, Lewis, tau Yih, Rocktäschel, Riedel, and Kiela}]{lewis2021retrievalaugmented}
Patrick Lewis, Ethan Perez, Aleksandra Piktus, Fabio Petroni, Vladimir Karpukhin, Naman Goyal, Heinrich Küttler, Mike Lewis, Wen tau Yih, Tim Rocktäschel, Sebastian Riedel, and Douwe Kiela. 2021.
\newblock \href {https://arxiv.org/abs/2005.11401} {Retrieval-augmented generation for knowledge-intensive nlp tasks}.
\newblock \emph{Preprint}, arXiv:2005.11401.

\bibitem[{Li et~al.(2024{\natexlab{a}})Li, Haider, and Callison-Burch}]{li-etal-2024-land}
Bryan Li, Samar Haider, and Chris Callison-Burch. 2024{\natexlab{a}}.
\newblock \href {https://doi.org/10.18653/v1/2024.naacl-long.213} {This land is {Your, My} land: Evaluating geopolitical bias in language models through territorial disputes}.
\newblock In \emph{Proceedings of the 2024 Conference of the North American Chapter of the Association for Computational Linguistics: Human Language Technologies (Volume 1: Long Papers)}, pages 3855--3871, Mexico City, Mexico. Association for Computational Linguistics.

\bibitem[{Li et~al.(2024{\natexlab{b}})Li, Shi, Liu, Yang, Payani, Liu, and Du}]{li2024quantifyingmultilingualperformancelarge}
Zihao Li, Yucheng Shi, Zirui Liu, Fan Yang, Ali Payani, Ninghao Liu, and Mengnan Du. 2024{\natexlab{b}}.
\newblock \href {https://arxiv.org/abs/2404.11553} {Quantifying multilingual performance of large language models across languages}.
\newblock \emph{Preprint}, arXiv:2404.11553.

\bibitem[{Lück et~al.(2018)Lück, Wessler, Wozniak, and Lycarião}]{doi:10.1177/1464884916680372}
Julia Lück, Hartmut Wessler, Antal Wozniak, and Diógenes Lycarião. 2018.
\newblock \href {https://doi.org/10.1177/1464884916680372} {Counterbalancing global media frames with nationally colored narratives: A comparative study of news narratives and news framing in the climate change coverage of five countries}.
\newblock \emph{Journalism}, 19(12):1635--1656.

\bibitem[{Mehdad et~al.(2012)Mehdad, Negri, and Federico}]{mehdad2012detecting}
Yashar Mehdad, Matteo Negri, and Marcello Federico. 2012.
\newblock Detecting semantic equivalence and information disparity in cross-lingual documents.
\newblock In \emph{Proceedings of the 50th Annual Meeting of the Association for Computational Linguistics (Volume 2: Short Papers)}, pages 120--124.

\bibitem[{Nair et~al.(2022)Nair, Yang, Lawrie, Duh, McNamee, Murray, Mayfield, and Oard}]{10.1007/978-3-030-99736-6_26}
Suraj Nair, Eugene Yang, Dawn Lawrie, Kevin Duh, Paul McNamee, Kenton Murray, James Mayfield, and Douglas~W. Oard. 2022.
\newblock \href {https://doi.org/10.1007/978-3-030-99736-6_26} {Transfer learning approaches for building cross-language dense retrieval models}.
\newblock In \emph{Advances in Information Retrieval: 44th European Conference on IR Research, ECIR 2022, Stavanger, Norway, April 10–14, 2022, Proceedings, Part I}, page 382–396, Berlin, Heidelberg. Springer-Verlag.

\bibitem[{OpenAI(2024)}]{openai2024gpt4technicalreport}
OpenAI. 2024.
\newblock \href {https://arxiv.org/abs/2303.08774} {Gpt-4 technical report}.
\newblock \emph{Preprint}, arXiv:2303.08774.

\bibitem[{Parton et~al.(2008)Parton, McKeown, Allan, and Henestroza}]{parton2008simultaneous}
Kristen Parton, Kathleen~R McKeown, James Allan, and Enrique Henestroza. 2008.
\newblock Simultaneous multilingual search for translingual information retrieval.
\newblock In \emph{Proceedings of the 17th ACM conference on Information and knowledge management}, pages 719--728.

\bibitem[{Phillipson(2018)}]{Phillipson2018LinguisticI}
Robert Phillipson. 2018.
\newblock \href {https://api.semanticscholar.org/CorpusID:239523960} {Linguistic imperialism}.
\newblock \emph{The Encyclopedia of Applied Linguistics}.

\bibitem[{Qin et~al.(2025)Qin, Chen, Zhou, Chen, Li, Liao, Li, Che, and Yu}]{QIN2025101118}
Libo Qin, Qiguang Chen, Yuhang Zhou, Zhi Chen, Yinghui Li, Lizi Liao, Min Li, Wanxiang Che, and Philip~S. Yu. 2025.
\newblock \href {https://doi.org/10.1016/j.patter.2024.101118} {A survey of multilingual large language models}.
\newblock \emph{Patterns}, 6(1):101118.

\bibitem[{Sharma et~al.(2024)Sharma, Liao, and Xiao}]{10.1145/3613904.3642459}
Nikhil Sharma, Q.~Vera Liao, and Ziang Xiao. 2024.
\newblock \href {https://doi.org/10.1145/3613904.3642459} {Generative echo chamber? effect of llm-powered search systems on diverse information seeking}.
\newblock In \emph{Proceedings of the CHI Conference on Human Factors in Computing Systems}, CHI '24, New York, NY, USA. Association for Computing Machinery.

\bibitem[{Sittar et~al.(2022)Sittar, Leban, Fortuna, and Grobelnik}]{sittar2021analysis}
Aisha Sittar, Gregor Leban, Blaz Fortuna, and Marko Grobelnik. 2022.
\newblock Analysis of information cascading and propagation barriers across distinctive news events.
\newblock \emph{Journal of Intelligent Information Systems}, 58:119--152.

\bibitem[{Verma et~al.(2022)Verma, Mujumdar, Wang, Choudhury, and Kumar}]{Verma_Mujumdar_Wang_Choudhury_Kumar_2022}
Gaurav Verma, Rohit Mujumdar, Zijie~J. Wang, Munmun~De Choudhury, and Srijan Kumar. 2022.
\newblock \href {https://doi.org/10.1609/icwsm.v16i1.19356} {Overcoming language disparity in online content classification with multimodal learning}.
\newblock \emph{Proceedings of the International AAAI Conference on Web and Social Media}, 16(1):1040--1051.

\bibitem[{Watts et~al.(2024)Watts, Gumma, Yadavalli, Seshadri, Swaminathan, and Sitaram}]{watts-etal-2024-pariksha}
Ishaan Watts, Varun Gumma, Aditya Yadavalli, Vivek Seshadri, Manohar Swaminathan, and Sunayana Sitaram. 2024.
\newblock \href {https://doi.org/10.18653/v1/2024.emnlp-main.451} {{PARIKSHA}: A large-scale investigation of human-{LLM} evaluator agreement on multilingual and multi-cultural data}.
\newblock In \emph{Proceedings of the 2024 Conference on Empirical Methods in Natural Language Processing}, pages 7900--7932, Miami, Florida, USA. Association for Computational Linguistics.

\bibitem[{White(1990)}]{white1990content}
Hayden White. 1990.
\newblock \emph{The content of the form: Narrative discourse and historical representation}.
\newblock JHU Press.

\bibitem[{Wu et~al.(2024)Wu, Ren, and Verberne}]{wu2024limitscrosslingualdensepassage}
Jie Wu, Zhaochun Ren, and Suzan Verberne. 2024.
\newblock \href {https://arxiv.org/abs/2408.11942} {What are the limits of cross-lingual dense passage retrieval for low-resource languages?}
\newblock \emph{Preprint}, arXiv:2408.11942.

\bibitem[{Xie et~al.(2024)Xie, Zhang, Chen, Lou, and Su}]{xie2024adaptive}
Jian Xie, Kai Zhang, Jiangjie Chen, Renze Lou, and Yu~Su. 2024.
\newblock \href {https://openreview.net/forum?id=auKAUJZMO6} {Adaptive chameleon or stubborn sloth: Revealing the behavior of large language models in knowledge conflicts}.
\newblock In \emph{The Twelfth International Conference on Learning Representations}.

\bibitem[{Xu et~al.(2024)Xu, Qi, Guo, Wang, Wang, Zhang, and Xu}]{xu-etal-2024-knowledge-conflicts}
Rongwu Xu, Zehan Qi, Zhijiang Guo, Cunxiang Wang, Hongru Wang, Yue Zhang, and Wei Xu. 2024.
\newblock \href {https://doi.org/10.18653/v1/2024.emnlp-main.486} {Knowledge conflicts for {LLM}s: A survey}.
\newblock In \emph{Proceedings of the 2024 Conference on Empirical Methods in Natural Language Processing}, pages 8541--8565, Miami, Florida, USA. Association for Computational Linguistics.

\bibitem[{Yang et~al.(2024)Yang, Lawrie, Mayfield, Oard, and Miller}]{10.1007/978-3-031-56060-6_4}
Eugene Yang, Dawn Lawrie, James Mayfield, Douglas~W. Oard, and Scott Miller. 2024.
\newblock \href {https://doi.org/10.1007/978-3-031-56060-6_4} {Translate-distill: Learning cross-language dense retrieval by translation and distillation}.
\newblock In \emph{Advances in Information Retrieval: 46th European Conference on Information Retrieval, ECIR 2024, Glasgow, UK, March 24–28, 2024, Proceedings, Part II}, page 50–65, Berlin, Heidelberg. Springer-Verlag.

\bibitem[{Yang et~al.(1998)Yang, Carbonell, Brown, and Frederking}]{YANG1998323}
Yiming Yang, Jaime~G. Carbonell, Ralf~D. Brown, and Robert~E. Frederking. 1998.
\newblock \href {https://doi.org/10.1016/S0004-3702(98)00063-0} {Translingual information retrieval: learning from bilingual corpora}.
\newblock \emph{Artificial Intelligence}, 103(1):323--345.
\newblock Artificial Intelligence 40 years later.

\bibitem[{Yu et~al.(2022)Yu, Chatterjee, Asai, Hu, and Choi}]{yu-etal-2022-beyond}
Xinyan Yu, Trina Chatterjee, Akari Asai, Junjie Hu, and Eunsol Choi. 2022.
\newblock \href {https://doi.org/10.18653/v1/2022.findings-emnlp.273} {Beyond counting datasets: A survey of multilingual dataset construction and necessary resources}.
\newblock In \emph{Findings of the Association for Computational Linguistics: EMNLP 2022}, pages 3725--3743, Abu Dhabi, United Arab Emirates. Association for Computational Linguistics.

\bibitem[{Zhang et~al.(2020)Zhang, Karakos, Hartmann, Srivastava, Tarlin, Akodes, Gouda, Bathool, Zhao, Jiang, Schwartz, and Makhoul}]{zhang-etal-2020-2019}
Le~Zhang, Damianos Karakos, William Hartmann, Manaj Srivastava, Lee Tarlin, David Akodes, Sanjay~Krishna Gouda, Numra Bathool, Lingjun Zhao, Zhuolin Jiang, Richard Schwartz, and John Makhoul. 2020.
\newblock \href {https://aclanthology.org/2020.clssts-1.8} {The 2019 {BBN} cross-lingual information retrieval system}.
\newblock In \emph{Proceedings of the workshop on Cross-Language Search and Summarization of Text and Speech (CLSSTS2020)}, pages 44--51, Marseille, France. European Language Resources Association.

\bibitem[{Zhang* et~al.(2020)Zhang*, Kishore*, Wu*, Weinberger, and Artzi}]{bert-score}
Tianyi Zhang*, Varsha Kishore*, Felix Wu*, Kilian~Q. Weinberger, and Yoav Artzi. 2020.
\newblock \href {https://openreview.net/forum?id=SkeHuCVFDr} {Bertscore: Evaluating text generation with bert}.
\newblock In \emph{International Conference on Learning Representations}.

\bibitem[{Zhang et~al.(2023)Zhang, Ogueji, Ma, and Lin}]{10.1145/3613447}
Xinyu Zhang, Kelechi Ogueji, Xueguang Ma, and Jimmy Lin. 2023.
\newblock \href {https://doi.org/10.1145/3613447} {Toward best practices for training multilingual dense retrieval models}.
\newblock \emph{ACM Trans. Inf. Syst.}, 42(2).

\bibitem[{Zhang et~al.(2011)Zhang, Wang, Wang, Xu, Xu, and Fan}]{zhang2011diversifying}
Yaoyun Zhang, Xiaolong Wang, Xuan Wang, Ruifeng Xu, Jun Xu, and Shixi Fan. 2011.
\newblock Diversifying information needs in results of question retrieval.
\newblock In \emph{Proceedings of 5th International Joint Conference on Natural Language Processing}, pages 1432--1436.

\end{thebibliography}
\clearpage

\appendix

\section{Appendix}
\subsection{Types of information conflicts}
\label{appendix:information conflicts}
Our dataset supports two common information conflicts: facts and opinions. We considered the following scenarios for each type of query:

\paragraph{Discrepancies in facts:} Documents about the same event in different languages often focus on different or conflicting facts due to cultural differences and historical narratives. 
    \begin{itemize}
        \item \textbf{Shared Facts}: same facts shared across different languages.
        \item \textbf{Contradictory Facts}: facts that contradict each other when in different languages.
        \item \textbf{Exclusive Facts}: facts that are exclusive to documents in a certain language.
    \end{itemize}

\paragraph{Diverse opinions and perspectives:} Documents, especially on complex geopolitical issues, may talk about the same event but with diverse perspectives and opinions in different languages.
\begin{itemize}
        \item \textbf{Neutral}: Documents that share a neutral perspective and a balanced view of the topic.
        \item \textbf{Opinionated}: Documents share an opinionated view about a topic, supporting the side that speaks the language.
    \end{itemize}

\subsection{Synthetic dataset setup}
\label{appendix: synthetic dataset setup}
An overview of the process of creating our dataset is illustrated in figure \ref{fig:Dataset-creation}. We describe the dataset creation steps in detail below:

\paragraph{Core facts:} To start, we created two variants for each premise: ``original'' and ``alternate''. We manually created a set of 10 core facts about the premise for the original set. For the alternate set, we edit these core facts based on the setup of the premise in Sec. ~\ref{sec: premise setup}. 
 
\paragraph{Extended story facts:} Since the festival premise by our design goals contains more short-form factual information instead of diverse perspectives, to mimic a more realistic setting, we expanded each core facts using OpenAI GPT-4-turbo model where we added a consistency check during the expansions such that none of the expansions contradict any of the core facts. The result was 5 extensions for each core fact, making a total of 50 story facts for the original document set and 50 for the alternate document set on each premise.

\paragraph{Final Documents:} Based on these extended story facts for the festival premises, we created a total of 14 documents, 7 for each document set in the festival setting using OpenAI GPT-4-turbo with temperature = 1. We set temperature = 1 to make the model's responses diverse. Of these 7 documents, 2 were long documents, consisting of a subset of 4 story facts in a story format, and 5 were short documents, consisting of a subset of 2 story facts each, in a news format mimicking articles in the real world. The average length of these documents was 484.85 words.

\paragraph{Manipulation Check:} We verified that the dataset is not in LLM's parametric memory by prompting each evaluated model with ``Tell me about {{named entity}}'' with temperature = 0. We ensured that the model did not provide any meaningful answers for each named entity (places, objects, names, etc.). We also performed a Google search of the named entities and ensured that no article about the named entities existed as of June 2024.

\paragraph{Multilingual Dataset}

To make our dataset multilingual, we translated documents and queries with translate-shell, a Python package that uses Google Translate at the backend, from English to the target language (e.g., de, hi, zh, hi). For the documents in the festival setting, we repeat the document creation step for each language, where we create a document set in English before translating it into the target language to mimic a real-world setting where the data is not entirely parallel.

\subsection{Cost of experiments}
We had 27 factual + 16 opinion-based queries. For all queries, we had 2 contexts: alternate and original. We had all permutations and combinations of these contexts across the 5 target languages, making it 5$\times$5 contexts for each premise (festival and war). We query each context with their respective query pool across all languages. This makes the total number of requests (27 factual queries + 16 opinion-based queries)  $\times$ 5 query-languages $\times$ 5 original context languages $\times$ 5 alternate context languages. Each input context was around 1000 tokens and 526 output tokens. This results in a total of 5,375,000 input tokens and 2,872,250 output tokens per model.
The cost of GPT-4o is \$5.00 / 1M tokens for inputs \$15.00 /1M tokens for outputs. This made the total cost of the experiments in the generation phase for OpenAI GPT-4o \$680.

\subsection{Prompt used in dataset creation}
During our dataset generation process, we use two OpenAI calls: one to expand the core facts and one to verify their consistency (filter). We repeat this for three iterations to ensure we have enough expansions after filtering.
We use the system prompt below with temperature t = 1 to expand the story. For the user prompt, we provide the model with JSON string of core facts.
\begin{tcolorbox}[fonttitle=\small\fon{pbk}\bfseries,
fontupper=\scriptsize\sffamily,
fontlower=\fon{put},
enhanced,
left=2pt, right=2pt, top=2pt, bottom=2pt,
title=Core Facts Expansion System Prompt]
\begin{lstlisting}[language=story]
You are a story points generator. Given a set of core points generate new point expansions for each of the core points such that it tangentially expands the story. The new points must be in the same setting but can talk about a different subtopic not connected to the main topic. The new points should not contradict any of the core points. You must come up with at least 4 new points for each core point. The output must be formatted in the following JSON structure:\n{\n\"CF_1\": {\n\"CF\": core fact,\n\"EF\": [expansions]\n},\n\"CF_2\": {\n\"CF\": core fact,\n\"EF\": [expansions]\n}...\n}
\end{lstlisting}
\end{tcolorbox}

We use the prompt below with temperature t = 0 to verify consistency.
\begin{tcolorbox}[fonttitle=\small\fon{pbk}\bfseries,
fontupper=\scriptsize\sffamily,
fontlower=\fon{put},
enhanced,
left=2pt, right=2pt, top=2pt, bottom=2pt,
title=Expanded Facts Verification System Prompt]
\begin{lstlisting}[language=story]
You are a story points filter. Given a set of core points and a new expansion, determine if the expansion contradicts any of the core points. If it does, output 'YES'. Otherwise, output 'NO'.
\end{lstlisting}
\end{tcolorbox}
\begin{tcolorbox}[fonttitle=\small\fon{pbk}\bfseries,
fontupper=\scriptsize\sffamily,
fontlower=\fon{put},
enhanced,
left=2pt, right=2pt, top=2pt, bottom=2pt,
title=Expanded Facts Verification User Prompt]
\begin{lstlisting}[language=story]
CORE POINTS:\n" + core_points_str + "\nEXPANSION:\n" + exp + "\nDoes this expansion contradict any of the core points? YES/NO
\end{lstlisting}
\end{tcolorbox}

To convert the expanded facts into documents, we use the following prompts:
\begin{tcolorbox}[fonttitle=\small\fon{pbk}\bfseries,
fontupper=\scriptsize\sffamily,
fontlower=\fon{put},
enhanced,
left=2pt, right=2pt, top=2pt, bottom=2pt,
title=News Document Creation System Prompt]
\begin{lstlisting}[language=story]
You are a news article generator. Given a set of story points, generate a news article that connects all the story points. The news must be coherent and should not contradict any of the story points. The output must be a piece of news that connects all the story points without adding any new story points. The output must be in "+ code_to_lang[language] + "."
\end{lstlisting}
\end{tcolorbox}

\begin{tcolorbox}[fonttitle=\small\fon{pbk}\bfseries,
fontupper=\scriptsize\sffamily,
fontlower=\fon{put},
enhanced,
left=2pt, right=2pt, top=2pt, bottom=2pt,
title=Story Document Creation System Prompt]
\begin{lstlisting}[language=story]
You are a story generator. Given a set of story points, generate a story that connects all the story points. The story must be coherent and should not contradict any of the story points. The output must be a story that connects all the story points without adding any new story points.
\end{lstlisting}
\end{tcolorbox}

To verify the story covers all the facts, we use another OpenAI call with the following prompt:
\begin{tcolorbox}[fonttitle=\small\fon{pbk}\bfseries,
fontupper=\scriptsize\sffamily,
fontlower=\fon{put},
enhanced,
left=2pt, right=2pt, top=2pt, bottom=2pt,
title=Story Verification System Prompt]
\begin{lstlisting}[language=story]
You are a story verifier. Given a STORY_POINTS and corresponding STORY, verify if the STORY contains all the STORY_POINTS. If the STORY contains all the STORY_POINTS, return 'True'. Otherwise, return 'False'. YOU MUST ONLY OUTPUT 'True' or 'False'."
\end{lstlisting}
\end{tcolorbox}

\begin{tcolorbox}[fonttitle=\small\fon{pbk}\bfseries,
fontupper=\scriptsize\sffamily,
fontlower=\fon{put},
enhanced,
left=2pt, right=2pt, top=2pt, bottom=2pt,
title=Story Verification User Prompt]
\begin{lstlisting}[language=story]
STORY_POINTS: " + json.dumps(story_points) + "\n" + "STORY: " + output_en + "\n Does the STORY contain all the STORY_POINTS? True or False"\end{lstlisting}
\end{tcolorbox}

\subsection{Retrieval Preference by Model}
\label{sec: appendix-retrieval}
\begin{figure}[H]
\includegraphics[width=0.63\linewidth]{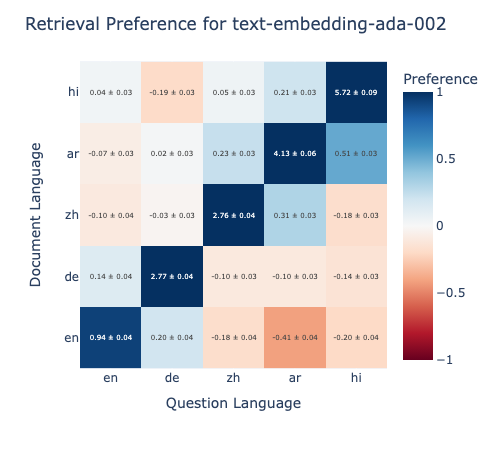}
\centering
\caption{The figure shows the aggregate language retrieval score for ada-002 model. }
\end{figure}

\begin{figure}[H]
\includegraphics[width=0.63\linewidth]{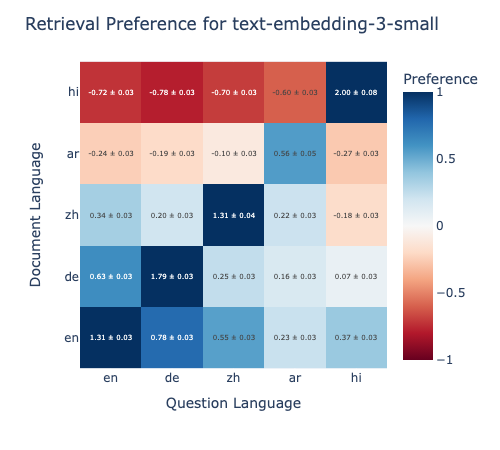}
\centering
\caption{The figure shows the aggregate language retrieval score for embedding-3-small model. }
\end{figure}

\begin{figure}[H]
\includegraphics[width=0.63\linewidth]{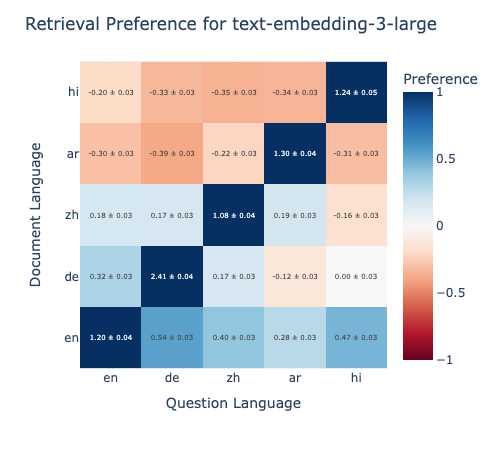}
\centering
\caption{The figure shows the aggregate language retrieval score for embedding-3-large model. }
\end{figure}

\begin{figure}[H]
\includegraphics[width=0.63\linewidth]{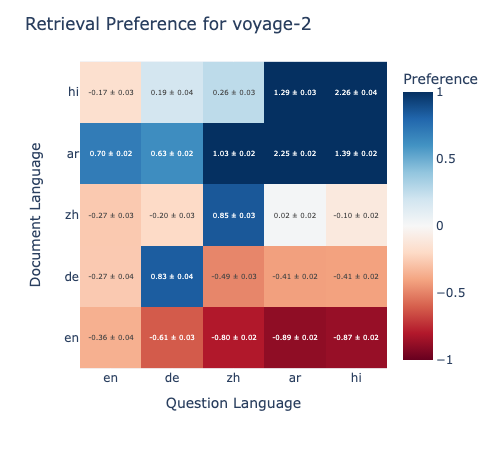}
\centering
\caption{The figure shows the aggregate language retrieval score for voyage-2 model. }
\end{figure}

\begin{figure}[H]
\includegraphics[width=0.63\linewidth]{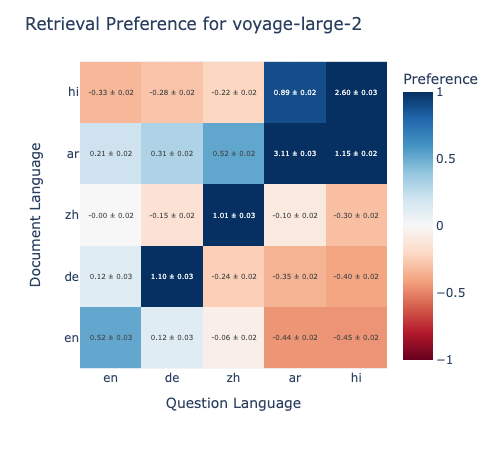}
\centering
\caption{The figure shows the aggregate language retrieval score for voyage-2-large model. }
\end{figure}

\begin{figure}[H]
\includegraphics[width=0.63\linewidth]{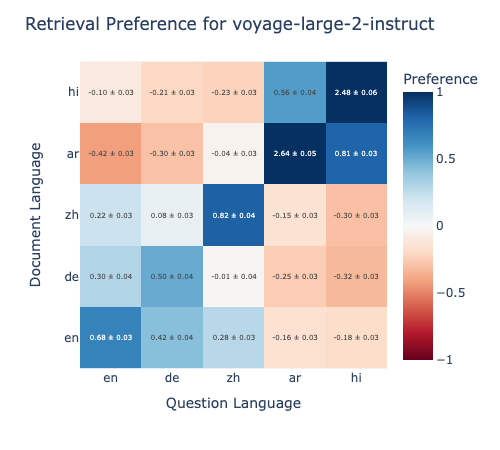}
\centering
\caption{The figure shows the aggregate language retrieval score for voyage-2-large-instruct model. }
\end{figure}

\begin{figure}[H]
\includegraphics[width=0.63\linewidth]{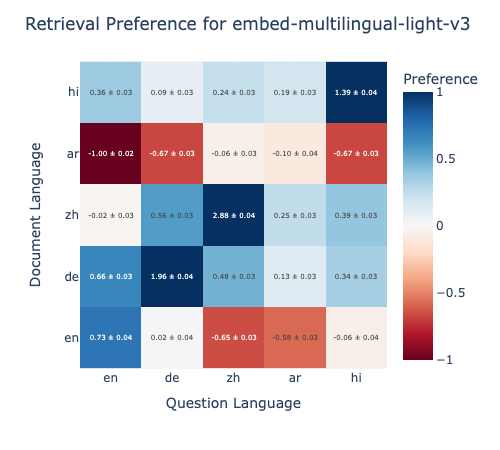}
\centering
\caption{The figure shows the aggregate language retrieval score for Cohere's multilingual v3 light model. }
\end{figure}

\begin{figure}[h!]
\includegraphics[width=0.60\linewidth]{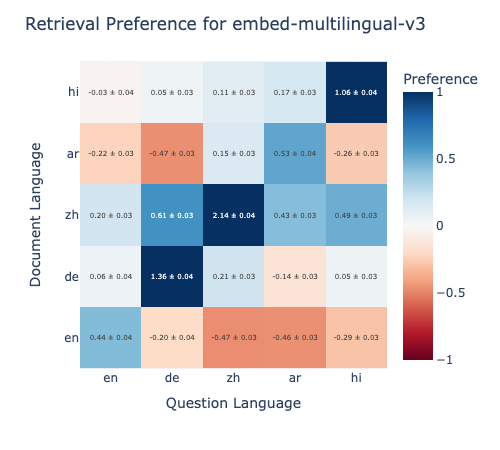}
\centering
\caption{The figure shows the aggregate language retrieval score for Cohere's multilingual v3 model. }
\end{figure}

\subsection{Generation Preference by Model}
\label{sec: appendix-generation}
\begin{figure}[H]
\includegraphics[width=0.60\linewidth]{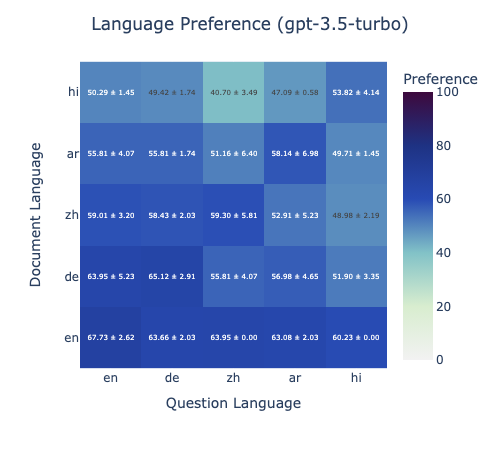}
\centering
\caption{The figure shows the aggregate source language preference for GPT-3.5-turbo model. }
\end{figure}

\begin{figure}[H]
\includegraphics[width=0.60\linewidth]{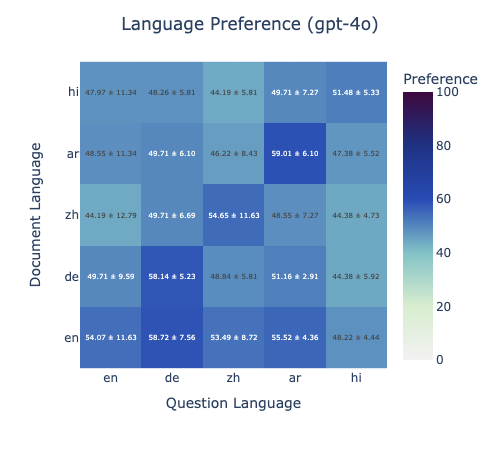}
\centering
\caption{The figure shows the aggregate source language preference for GPT-4o model. }
\end{figure}

\begin{figure}[H]
\includegraphics[width=0.60\linewidth]{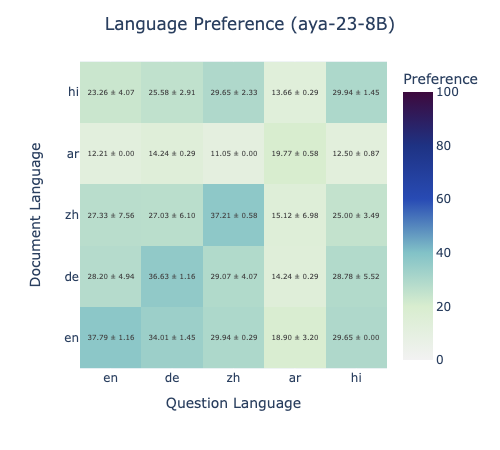}
\centering
\caption{The figure shows the aggregate source language preference for aya-23-8B model. }
\end{figure}

\begin{figure}[H]
\includegraphics[width=0.63\linewidth]{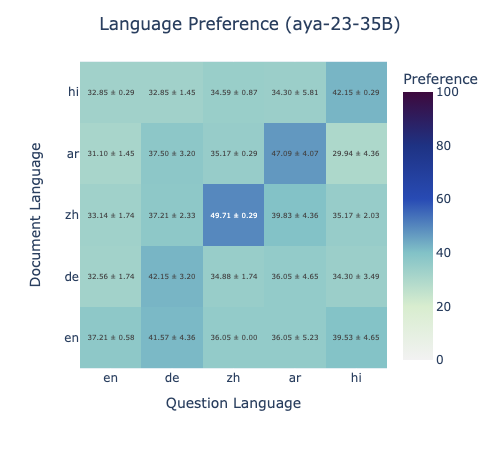}
\centering
\caption{The figure shows the aggregate source language preference for aya-23-35B model. }
\end{figure}

\begin{figure}[H]
\includegraphics[width=0.63\linewidth]{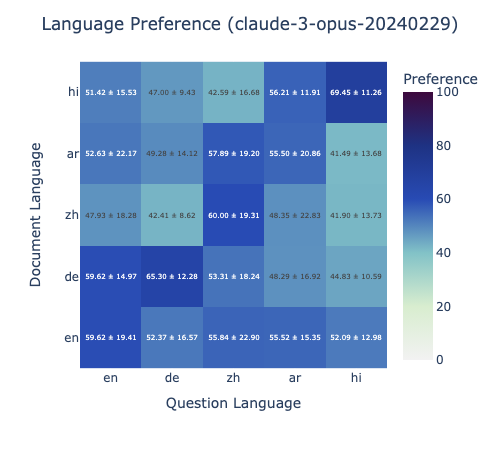}
\centering
\caption{The figure shows the aggregate source language preference for Claude-3-opous model. }
\end{figure}

\subsubsection{Generation Preference all Models}

\label{sec: appendix-all-none-both}
\begin{figure}[H]
\includegraphics[width=0.90\linewidth]{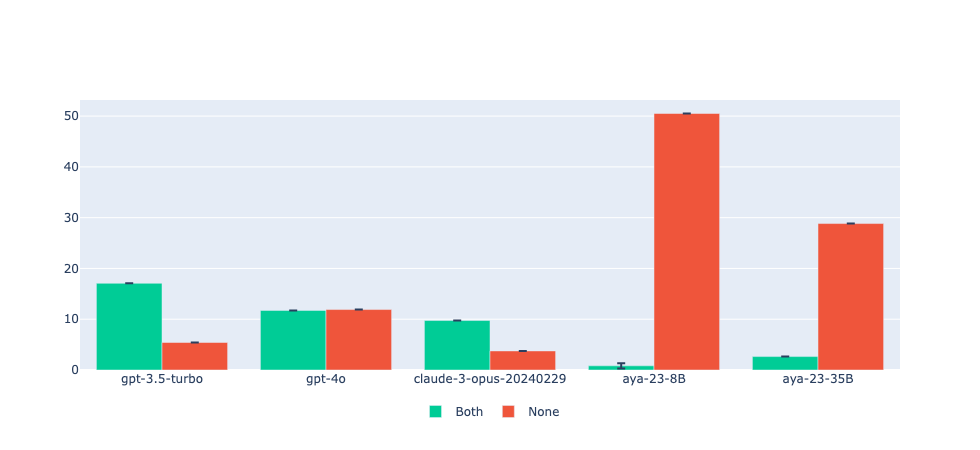}
\centering
\caption{The figure shows the percentage of \textit{Both} and \textit{None} answers by different models where in \textit{Both} answers, the model incorporated information from both in-context documents and in \textit{None} answers, the model cannot provide meaningful answers. }
\end{figure}

\subsubsection{Preference for \textit{Both} when both context documents are in Foreign Languages}
\label{sec: appendix-all-both}
\begin{figure}[H]
\includegraphics[width=0.55\linewidth]{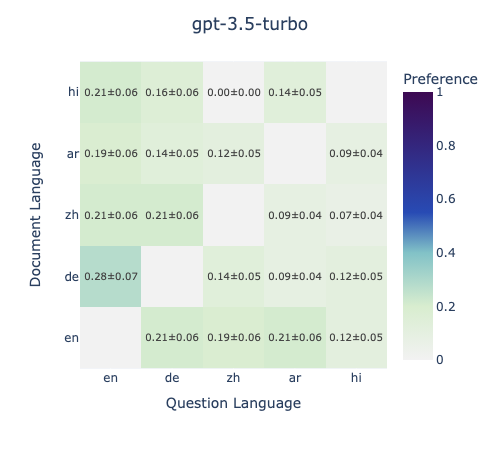}
\centering
\caption{The figure shows the preference for both documents in GPT-3.5-turbo model.}
\end{figure}

\begin{figure}[H]
\includegraphics[width=0.55\linewidth]{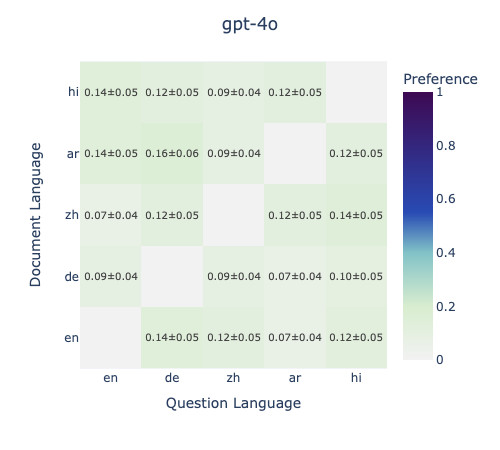}
\centering
\caption{The figure shows the preference for both documents in GPT-4o model. }
\end{figure}

\begin{figure}[H]
\includegraphics[width=0.55\linewidth]{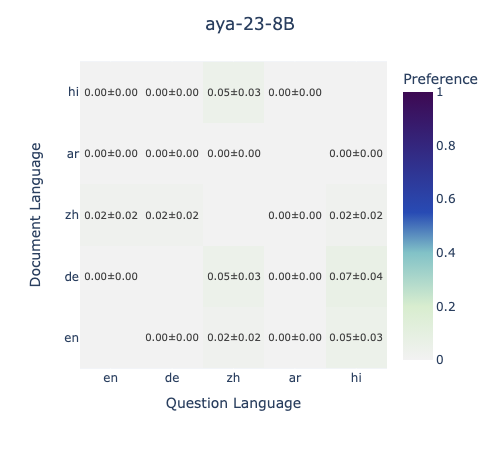}
\centering
\caption{The figure shows the preference for both documents in aya-23-8B model. }
\end{figure}

\begin{figure}[H]
\includegraphics[width=0.55\linewidth]{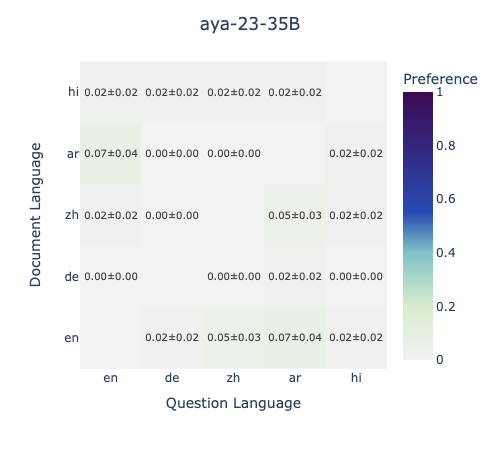}
\centering
\caption{The figure shows the preference for both documents in aya-23-35B model. }
\end{figure}

\begin{figure}[H]
\includegraphics[width=0.55\linewidth]{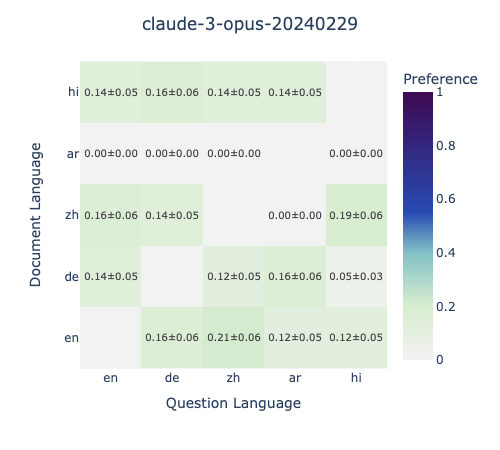}
\centering
\caption{The figure shows the preference for both documents in Claude-3-opous model. }
\end{figure}

\subsubsection{Preference for \textit{None} when both
context documents are in Foreign
Languages}
\label{sec: appendix-all-none}

\begin{figure}[H]
\includegraphics[width=0.55\linewidth]{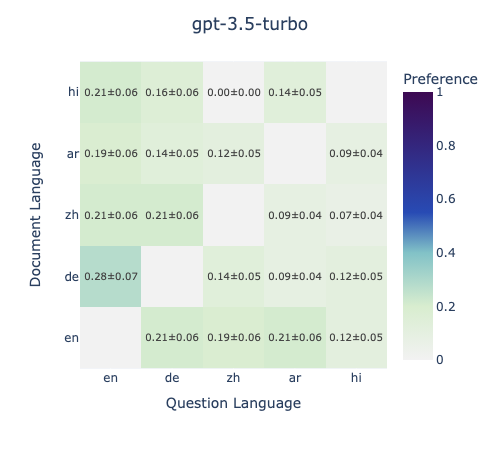}
\centering
\caption{The figure shows the preference for none of the documents for GPT-3.5-turbo model. }
\end{figure}

\begin{figure}[H]
\includegraphics[width=0.55\linewidth]{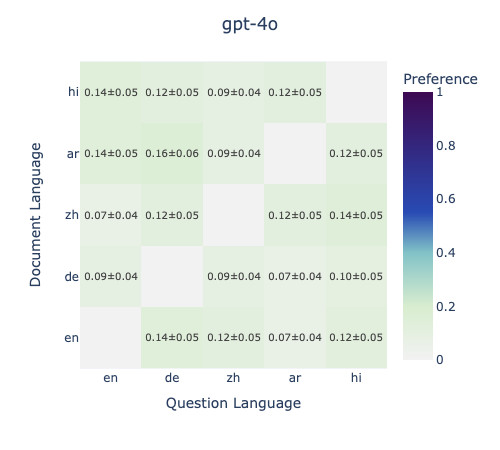}
\centering
\caption{The figure shows the preference for none of the documents for GPT-4o model. }
\end{figure}

\begin{figure}[H]
\includegraphics[width=0.55\linewidth]{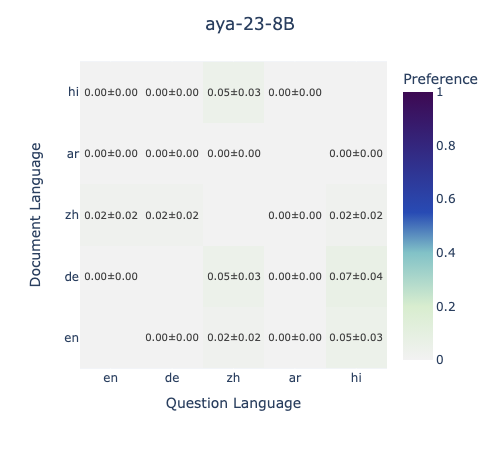}
\centering
\caption{The figure shows the preference for none of the documents for aya-23-8B model. }
\end{figure}

\begin{figure}[H]
\includegraphics[width=0.55\linewidth]{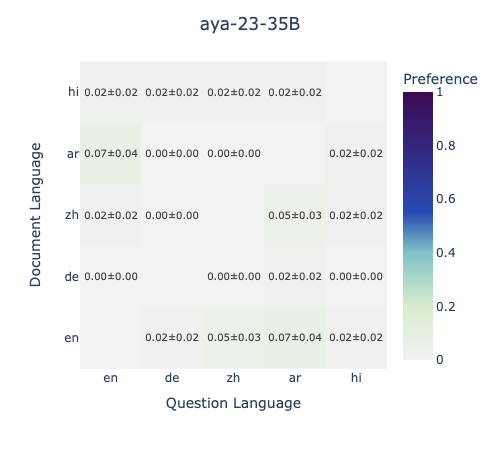}
\centering
\caption{The figure shows the preference for none of the documents for aya-23-35B model. }
\end{figure}

\begin{figure}[H]
\includegraphics[width=0.55\linewidth]{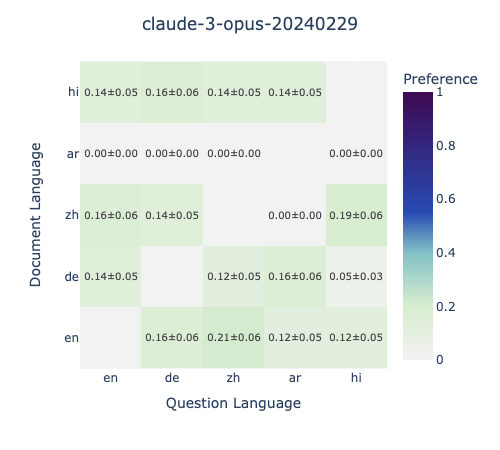}
\centering
\caption{The figure shows the preference for none of the documents for Claude-3-opous model. }
\end{figure}

\subsection{Additional Results}
\label{sec: appendix-new}
\subsubsection{Retrieval}
\begin{figure}[H]
\includegraphics[width=0.60\linewidth]{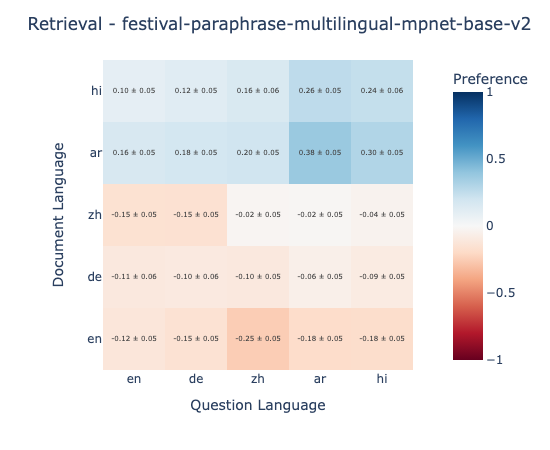}
\centering
\caption{The figure shows the aggregate language preferences during factual queries for the multilingual-mpnet-base-v2 model.}
\end{figure}

\begin{figure}[H]
\includegraphics[width=0.60\linewidth]{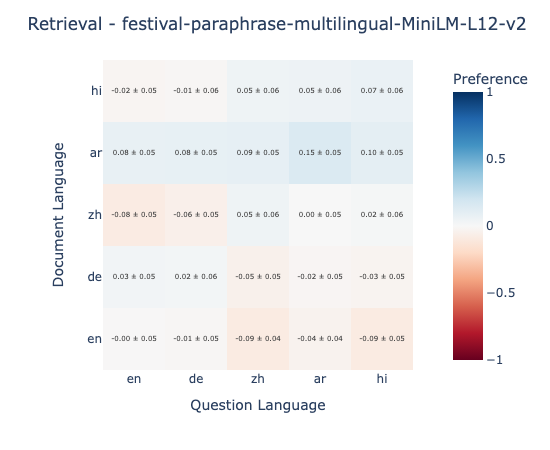}
\centering
\caption{The figure shows the aggregate language preferences during factual queries for multilingual MiniLM-L12-v2 model. }
\end{figure}

\begin{figure}[H]
\includegraphics[width=0.64\linewidth]{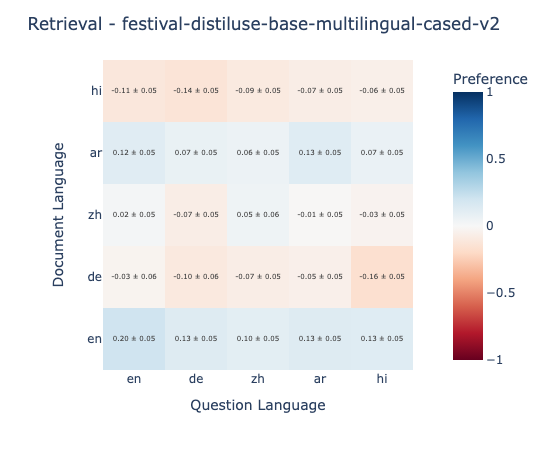}
\centering
\caption{The figure shows the aggregate language preferences during factual queries for distiluse-base-multilingual-cased model. }
\end{figure}

\begin{figure}[H]
\includegraphics[width=0.64\linewidth]{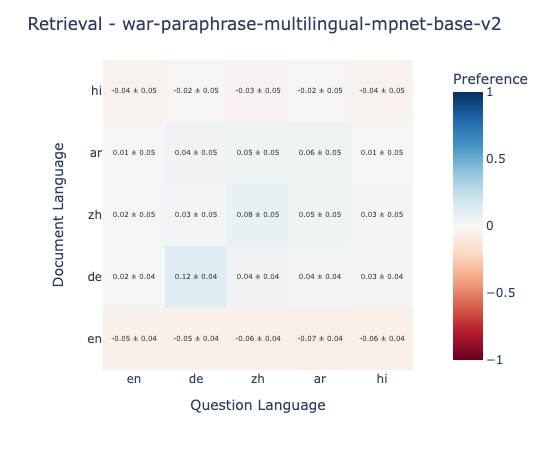}
\centering
\caption{The figure shows the aggregate language preferences during opinion queries for the (multilingual-mpnet-base-v2) model. }
\end{figure}

\begin{figure}[H]
\includegraphics[width=0.64\linewidth]{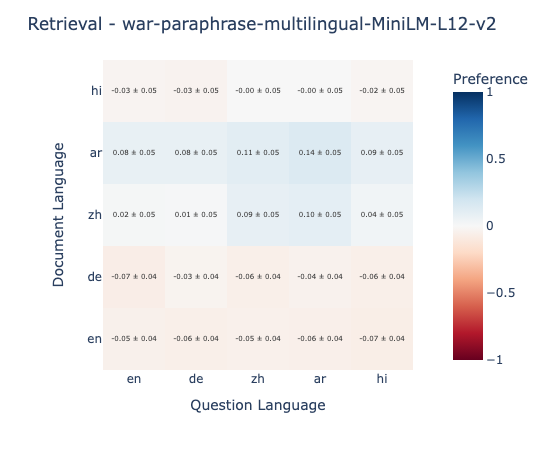}
\centering
\caption{The figure shows the aggregate language preferences during opinion queries for the multilingual MiniLM-L12-v2 model.}
\end{figure}

\begin{figure}[H]
\includegraphics[width=0.62\linewidth]{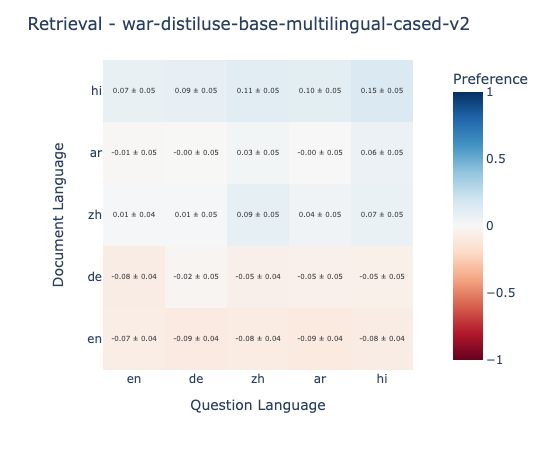}
\centering
\caption{The figure shows the aggregate language preferences during opinion queries for the distiluse-base-multilingual-cased model}
\end{figure}

\subsubsection{Generation}
\label{appx: generation-chinese}
\begin{table}[h!]
    \centering
    \small
    \begin{tabular}{c|c|c|c|c}
        query & en & zh & both & none\\ \hline
        en & \textbf{53.44} & 31.5 & 10.9 & 4.1 \\ 
        de & \textbf{55.8}  & 37.2 & 5.8 & 1.1 \\
        zh & \textbf{44.1} & 41.6 & 9.5 & 4.7 \\
        ar & \textbf{54.6} & 25.5 & 6.9 & 12.7 \\
        hi & \textbf{51.1} & 32.1 & 10.7 & 5.9 \\
    \end{tabular}
    \caption{We ran our test on moonshot-v1-8k, a model tailored for Chinese. We saw a preference for English language across all query languages, even when the query was in Chinese.}
    \label{tab:my_label}
\end{table}
\end{document}